\documentclass[10pt,twocolumn,letterpaper]{article}

\usepackage{wrapfig}

\usepackage{adjustbox}
\usepackage{diagbox}
\usepackage{subcaption}
\usepackage{graphicx}
\usepackage{multirow}
\usepackage{makecell}
\usepackage[table,dvipsnames]{xcolor}
\usepackage{mathrsfs}
\usepackage{pifont}
\usepackage{dsfont}

\newcommand{\cmark}{\ding{51}}%

\definecolor{customgreen}{RGB}{12, 180, 88}
\definecolor{pastelYellow}{RGB}{255, 239, 153}
\definecolor{pastelGreen}{RGB}{182, 225, 175}
\definecolor{pastelBlue}{RGB}{174, 198, 232}
\definecolor{F7E0D5}{RGB}{247,224,213}
\colorlet{Light}{White!0!F7E0D5}

\newlength\savewidth\newcommand\shline{\noalign{\global\savewidth\arrayrulewidth
\global\arrayrulewidth 1pt}\hline\noalign{\global\arrayrulewidth\savewidth}}
\newcommand{\tablestyle}[2]{\setlength{\tabcolsep}{#1}\renewcommand{\arraystretch}{#2}\centering\footnotesize}
\renewcommand{\paragraph}[1]{\vspace{1.25mm}\noindent\textbf{#1}}

\newcolumntype{x}[1]{>{\centering\arraybackslash}p{#1pt}}
\newcolumntype{y}[1]{>{\raggedright\arraybackslash}p{#1pt}}
\newcolumntype{z}[1]{>{\raggedleft\arraybackslash}p{#1pt}}

\usepackage[pagenumbers]{iccv} 

\definecolor{iccvblue}{rgb}{0.21,0.49,0.74}
\usepackage[pagebackref,breaklinks,colorlinks,allcolors=iccvblue]{hyperref}

\def\paperID{****} 
\def\confName{****}
\def\confYear{****}

\title{When Test-Time Adaptation Meets Self-Supervised Models}

\author{Jisu Han$^{1}$\thanks{Equal contribution.} \quad   Jihee Park$^{1*}$ \quad Dongyoon Han$^{2}$\quad  Wonjun Hwang$^{1}$\thanks{Corresponding author.}\\
$^1$Korea University \quad  $^2$NAVER AI Lab\\
}
\begin{document}
\maketitle

\begin{abstract}
Training on test-time data enables deep learning models to adapt to dynamic environmental changes, enhancing their practical applicability. Online adaptation from source to target domains is promising but it remains highly reliant on the performance of source pretrained model. In this paper, we investigate whether test-time adaptation (TTA) methods can continuously improve models trained via self-supervised learning (SSL) without relying on source pretraining. We introduce a self-supervised TTA protocol after observing that existing TTA approaches struggle when directly applied to self-supervised models with low accuracy on the source domain. Furthermore, we propose a collaborative learning framework that integrates SSL and TTA models, leveraging contrastive learning and knowledge distillation for stepwise representation refinement. We validate our method on diverse self-supervised models, including DINO, MoCo, and iBOT, across TTA benchmarks. Extensive experiments validate the effectiveness of our approach in SSL, showing that it achieves competitive performance even without source pretraining.
\end{abstract}

\section{Introduction}
Deep neural networks (DNNs) have achieved remarkable advancements across various fields~\cite{he2016resnet,dosovitskiy2021vit,chen2017deeplab,redmon2016yolo} of computer vision and are increasingly becoming a standard tool in the industry~\cite{wang2023yolov7,wu2024ptv3,kerbl3Dgaussians}. However, the issue of performance degradation due to domain shift~\cite{domainshift} between training and test datasets remains an unresolved challenge, even when distributional differences appear to be minimal~\cite{recht2018cifar}.
To address this challenge, Test-Time Training (TTT) introduces a new paradigm in domain adaptation by training at test-time to address distributional shifts between training and test data~\cite{sun2020ttt,liu2021tttplus,gandelsman2022tttmae}. 
Building on the principles of TTT, various protocols have been developed to extend its practicality. Test-Time Adaptation (TTA) further extends this idea by adapting a pretrained model to the test domain without requiring access to source data, addressing concerns related to privacy and memory constraints~\cite{DequanWangetal2021,zhang2022memo,niu2023sar,lee2024deyo}, and Continual Test-Time Adaptation (CTTA) extends TTA by assuming a continuously evolving test distribution, where the model adapts sequentially over time~\cite{Wangetal2022cotta,brahma2023petal,liu2023vida,hanranked}.

\begin{figure}[t]
\begin{center}
\includegraphics[width=0.95\linewidth]{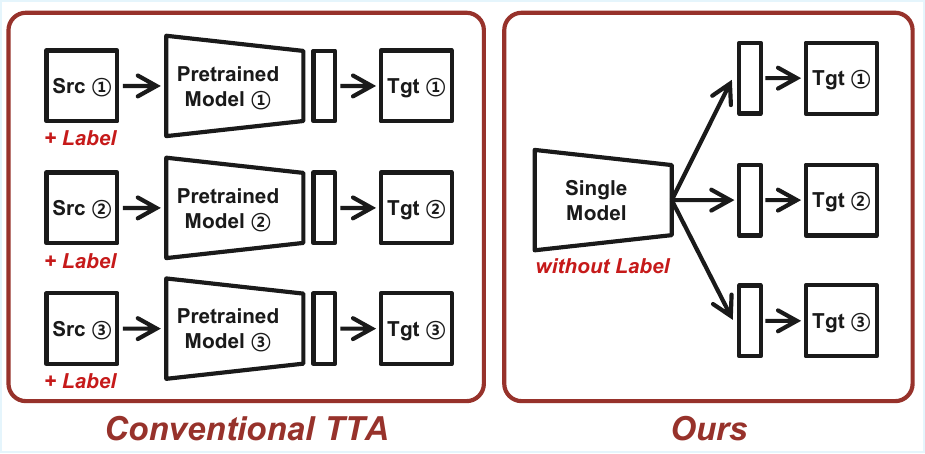}
\end{center}
\vspace{-4mm}
\caption{\textbf{Concept of Self-Supervised TTA.} Conventional TTA methods require a separate pretraining for each source domain, whereas our Self-Supervised TTA eliminates the need for source-specific pretraining by leveraging self-supervised learning.}
\label{fig:ssl_vs_sl}
\vspace{-2mm}
\end{figure}

\begin{table}[t]
    \centering
    \begin{adjustbox}{width=0.95\linewidth}
    \begin{tabular}{l  c c}
        \toprule
        \textbf{Source Model} & \textbf{ImageNet} & \textbf{CIFAR100} \\
        \midrule
        Source Pretraining & 1h8m23s$\times$300epochs & 9m7s$\times$200epochs  \\
        \rowcolor{Light!40}SSL w/ Prototype & 36m25s & 1m25s \\
        \rowcolor{Light!90}SSL w/ Prototype (Few-Shot)  & 1m56s &  7s \\
        \bottomrule
    \end{tabular}
    \end{adjustbox}
    \caption{\textbf{Training time comparison} between the source pretraining of the conventional TTA and our approach.
    }
    \vspace{-3.0mm}
    \label{tab:pretraining_time}
\end{table}

Despite many achievements of TTA, discussions on the pretraining model prepared using source data and corresponding labels have been limited. For example, as shown in~\cref{fig:ssl_vs_sl}, conventional TTA required a pretraining model trained on CIFAR10~\cite{krizhevsky2009cifar} to adapt to CIFAR10C (i.e., corruption set), but this model did not perform well on CIFAR100C. In other words, a separate pretraining model had to be prepared for each target domain. This limitation poses challenges in terms of practical efficiency and generality.
\begin{figure*}[t]
\begin{center}
    \begin{subfigure}[b]{0.32\linewidth}
        \centering
        \includegraphics[width=\linewidth]{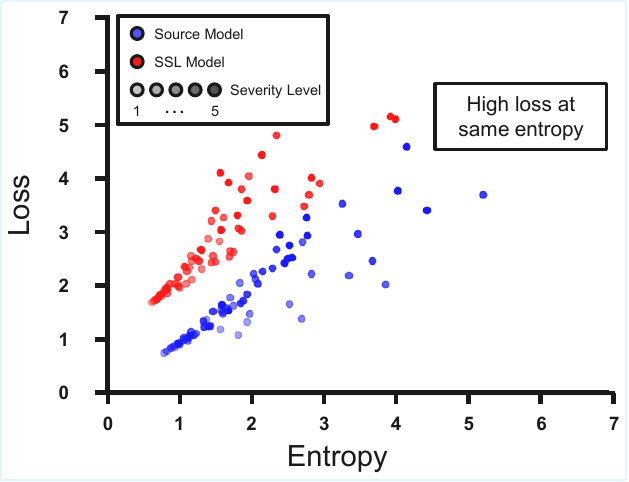}
        \caption{Failure of the EM Method in SSL model}
        \label{fig:analysis_1}
    \end{subfigure}
    \hfill
    \begin{subfigure}[b]{0.32\linewidth}
        \centering
        \includegraphics[width=\linewidth]{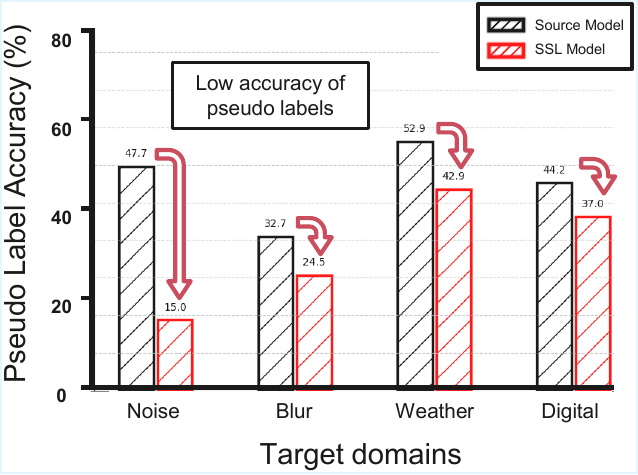}
        \caption{Failure of the CR Method in SSL model}
        \label{fig:analysis_2}
    \end{subfigure}
    \hfill
    \begin{subfigure}[b]{0.31\linewidth}
        \centering
        \includegraphics[width=\linewidth]{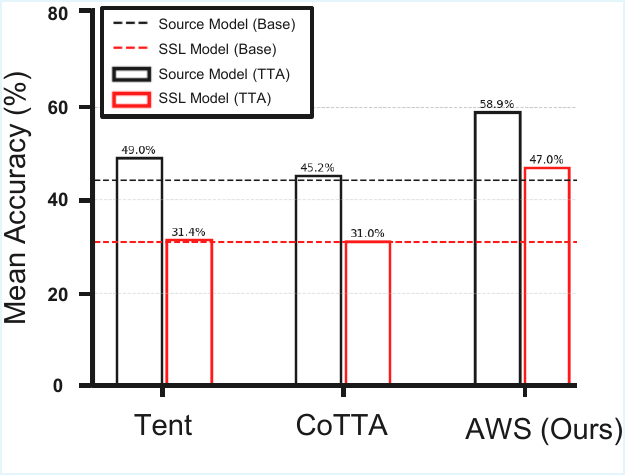}
        \caption{Performance comparison in SSL model}
        \label{fig:analysis_3}
    \end{subfigure}
\end{center}
\vspace{-5.0mm}
\caption{\textbf{Analysis of Self-Supervised models in Test-Time Adaptation.} (a) The relationship between entropy and loss for source pretrained and SSL models. SSL models tend to exhibit higher loss for the same entropy level and may decrease the entropy of incorrect predictions, thereby increasing the true risk.
(b) The accuracy of pseudo-labels for different target domains. SSL models generate pseudo labels with lower accuracy compared to source pre-trained models, which hinders performance improvement due to the propagation of inaccurate supervision signals.
(c) Comparison of accuracy across different TTA approaches. Our AWS achieves improved performance for the SSL model compared with  EM~\cite{DequanWangetal2021} and CR~\cite{Wangetal2022cotta} methods.}
\vspace{-1.0mm}
\label{fig:analysis}
\end{figure*}

Along with this, our study began with a simple question:
``\textit{Is the computational cost of pretraining the source model negligible compared to the adaptation process for unlabeled target data in TTA?}" 
We unveil the training time required for TTA methods using a pretrained source model in~\cref{tab:pretraining_time}, shedding light on the overlooked cost of source domain training and bringing it into the discussion. 
Optimizing the pretraining process of the source model is a practical matter, especially considering that labeled source data is often unavailable or prohibitively expensive to obtain.
A simple solution is to leverage the zero-shot performance of a self-supervised model trained through Self-Supervised Learning (SSL) on large-scale datasets~\cite{caron2021dino,chen2021moco,zhou2021ibot,cherti2023openclip,oquab2024dinov2}. This approach enhances generalization without requiring explicit supervision from the source domain, thereby mitigating the computational burden associated with pretraining while maintaining competitive adaptation performance in target domains. Specifically, we improve computational efficiency by designing a distance-based classifier that utilizes class prototypes obtained only through forward passes.

In this paper, we conduct an empirical investigation into the effectiveness of existing TTA approaches on self-supervised models without domain-specific knowledge and explore the feasibility of applying SSL for TTA. \cref{fig:analysis_1,fig:analysis_2} show that the primary TTA approaches, Entropy Minimization (EM)~\cite{DequanWangetal2021} and Consistency Regularization (CR)~\cite{Wangetal2022cotta}, are not readily applicable to SSL models. EM method minimizes predictive entropy based on the observation that lower entropy indicates higher model accuracy. While it has been demonstrated to be effective for conventional TTA, its applicability remains challenging in SSL models, where low entropy does not ensure accurate predictions.
Furthermore, CR approaches that leverage pseudo-labels to maintain predictive consistency also suffer from the inaccuracy of pseudo-labels based on the low domain accuracy of SSL models. 

Given that the SSL model does not seamlessly extend to TTA, we introduce a novel framework called Adapt Without Source pretraining (AWS). The proposed method consists of three key components. First, contrastive learning enhances the representation capability for both source and target domains. Second, knowledge distillation preserves the generalization ability of the initial SSL model. Third, mutual learning integrates the advantages of different predictions from the SSL and target models. \cref{fig:analysis_3} presents the TTA performance of a source model trained with supervised learning on the source domain and a self-supervised model, DINO~\cite{caron2021dino}. Compared to EM and CR approaches, which fail to enhance the performance of SSL models, our method demonstrates its effectiveness in improving TTA performance for SSL models. Notably, despite the initial performance gap on the target domain, our approach surpasses the source-pretrained model, highlighting the potential for advancing TTA using SSL models.
In summary, the main contributions of our work are as follows:
\begin{itemize}
    \item To the best of our knowledge, we are the first to highlight the issue of computational efficiency in the source training process for TTA. Motivated by this challenge, we propose a self-supervised test-time adaptation protocol.
    \item We investigate the potential of using self-supervised models in TTA and identify failure cases through an empirical analysis of prominent methodologies in TTA
    \item Beyond conventional TTA approaches that rely on the performance of a source model, we propose AWS, a novel method consisting of contrastive learning, knowledge distillation, and mutual learning. Our approach is extensively evaluated under standard TTA protocols and across various SSL models, demonstrating its effectiveness.
\end{itemize}

\begin{table*}[t]
    \centering
    \begin{adjustbox}{width=0.95\linewidth}
    \begin{tabular}{l c c c c c c}
        \toprule
        \textbf{Setting} & \textbf{Pretrain Loss} & \textbf{Source Data} & \textbf{Target Data} & \textbf{Data Distribution} & \textbf{Train Loss} & \textbf{Test Loss} \\
        \midrule
        Fine-Tuning & - & - & $x^t, y^t$ & -& $L(x^t, y^t)$ & - \\
        Domain Adaptation & - & $x^s, y^s$& $x^t$ &- & $L(x^s, y^s) + L(x^t, x^s)$ & - \\
        Source-Free Domain Adaptation & $L(x^s, y^s)$ & - & $x^t$ &-& $L(x^t)$ & -  \\
        Test-Time Training & - & $x^s, y^s$ & $x^t$ &stationary& $L(x^s, y^s) + L(x^t)$ & $L(x^t)$ \\
        Fully Test-Time Adaptation & $L(x^s, y^s)$ & - & $x^t$ &stationary & - & $L(x^t)$ \\
        Continual Test-Time Adaptation & $L(x^s, y^s)$ & - & $x^t$ &continually changing & - & $L(x^t)$ \\
        \rowcolor{Light}Self-Supervised Test-Time Adaptation & $L(x^u)$ & - & $x^t$ & continually changing & - & $L(x^t)$ \\
        \bottomrule
    \end{tabular}
    \end{adjustbox}
    \caption{\textbf{Comparison of different adaptation protocols.} 
    Existing protocols require training on source data during the adaptation or pretraining process. Self-Supervised Test-Time Adaptation is based on unlabeled data $x^u$, which is not the source domain, and does not involve training on the source data. For source domain, only a forward pass over full or few-shot is performed, without backpropagation.}
    \label{tab:adaptation}
\end{table*}

\section{Related Work}
\subsection{Test-Time Adaptation}
Distributional discrepancies between the source and target domains present a significant challenge during the deployment of DNNs~\cite{domainshift}, and TTT introduces a learning approach that operates during test time~\cite{sun2020ttt}. TTT mitigates domain shift by adopting supervised learning on the source domain and self-training on unlabeled target domain data~\cite{liu2021tttplus,gandelsman2022tttmae,osowiechi2024ncttt}. In contrast, TTA emphasizes the impracticality of accessing source domain data and instead proposes an adaptation strategy that is solely applied at test time using a source pretrained model~\cite{DequanWangetal2021}. The main solution for TTA is the EM-based approach~\cite{niu2022eata,niu2023sar,lee2024deyo,zhang2025come}. The EM approach updates only the normalization layer and filters out inaccurate samples from the observation that samples with low entropy perform relatively well. Moreover, CTTA proposes a solution to address scenarios involving continuous domain shifts~\cite{Wangetal2022cotta}. CR is a primary solution in CTTA and has gained prominence for its effectiveness in stabilizing adaptation over time~\cite{Wangetal2022cotta,brahma2023petal,liu2023vida,liu2024continual}. The CR approach utilizes a teacher-student framework~\cite{tarvainen2017meanteacher} that updates all model parameters, enabling gradual adaptation through Exponential Moving Average (EMA) update. By leveraging pseudo labels generated by an augmented teacher model, CR enforces consistency throughout the adaptation process.

\subsection{Self-Supervised Learning}
The training of increasingly deeper and more complex DNNs demands large amounts of data. However, the expensive cost of human annotation presents challenges for supervised learning. SSL has been proposed as an alternative, leveraging unlabeled data for various downstream tasks~\cite{oord2018representation,he2020momentum,chen2021moco,chen2020simple,caron2021dino,zhou2021ibot,oquab2024dinov2}. CPC~\cite{oord2018representation} introduces a representation learning approach based on probabilistic contrastive learning for future prediction. MoCo~\cite{he2020momentum} employs a memory bank and a momentum encoder to facilitate contrastive learning with a large and consistent set of negative samples. SimCLR~\cite{chen2020simple} leverages strong data augmentations and a contrastive loss to maximize similarity between augmented views of the same instance. DINO~\cite{caron2021dino} adopts a self-distillation and teacher-student framework with a momentum encoder. iBOT~\cite{zhou2021ibot} proposes an mask prediction-based SSL framework through masked image modeling.

In this paper, we empirically investigate the effectiveness of TTA strategies in practical scenarios where labels are unavailable during the source pretraining phase. Furthermore, we propose Self-Supervised TTA, which leverages an SSL model as the source model and integrates it into the TTA.
\begin{figure}[t]
\begin{center}
\includegraphics[width=0.97\linewidth]{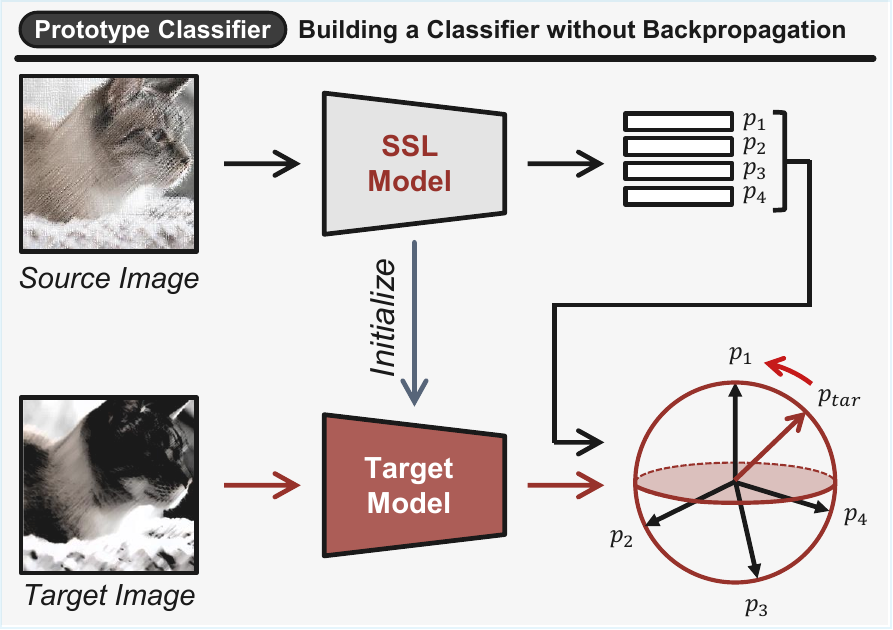}
\end{center}
\vspace{-4mm}
\caption{\textbf{A framework without source pretraining.} We construct a prototype classifier only through forward passes without a training process on the source domain.}
\label{fig:ssl}
\vspace{-2mm}
\end{figure}

\section{Self-Supervised Test-Time Adaptation}
\begin{figure*}[t]
\begin{center}
\includegraphics[width=0.99\linewidth]{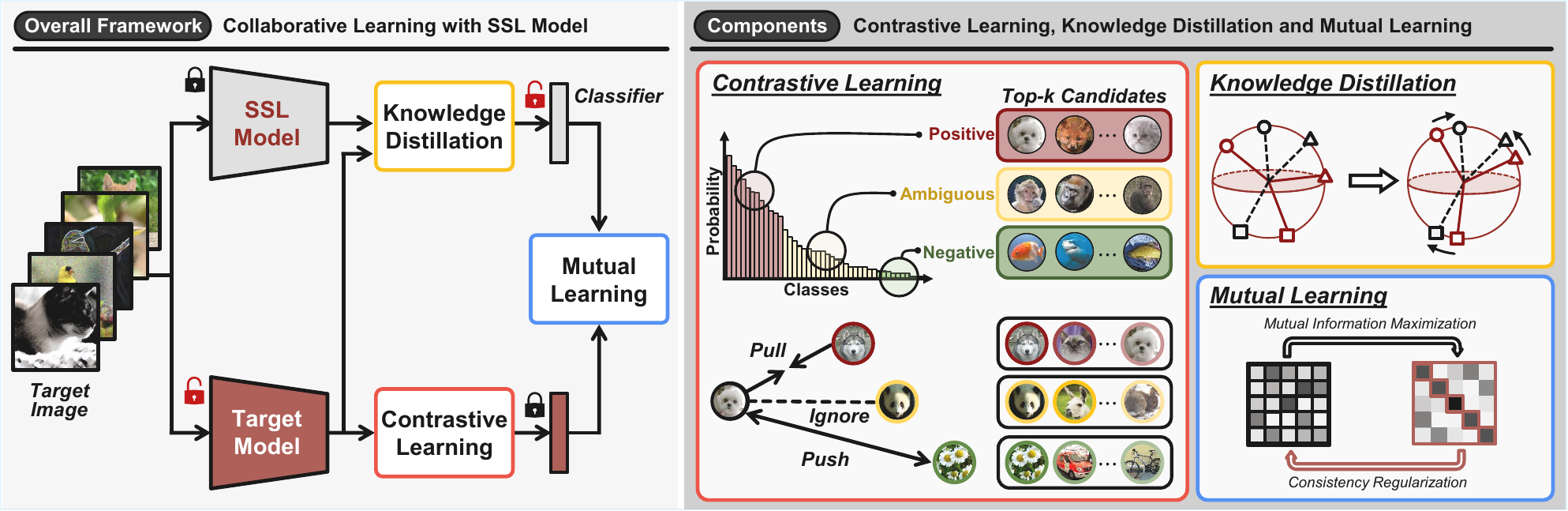}
\end{center}
\caption{\textbf{Overview of our AWS framework.} Contrastive learning refines representations by leveraging pseudo-labels while maintaining stability, knowledge distillation preserves generalization by aligning feature representations to mitigate overfitting under domain shifts, and mutual learning improves adaptation by integrating the generalization ability of the SSL model with the domain-specific knowledge of the target model through pseudo-labeling. }
\label{fig:overall}
\end{figure*}

\subsection{Protocol}
We briefly summarize the well-known adaption protocols for simple comparison in~\cref{tab:adaptation}, including the method replacing the source pre-training process in~\cref{fig:ssl} and the overview of our method is also illustrated in~\cref{fig:overall}.

\noindent\textbf{Source Model.}
Conventional TTA protocols~\cite{DequanWangetal2021,zhang2022memo,niu2023sar,Wangetal2022cotta,liu2023vida,liu2024continual} based on supervised learning of a source model \( g_s \circ f_s \) using labeled source domain data \( (x^s, y^s) \in \{\mathcal{X}^s,\mathcal{Y}^s\} \), where \( g_s \) and \( f_s \) represent the classifier and feature extractor of the source model, respectively. Instead of requiring pretraining on the source domain, we employ a self-supervised model \( f_{\text{ssl}} \) trained on an unlabeled data \( x^u \in \mathcal{X}^u \). We compute feature prototypes from either a subset or the entire source dataset to align the representation of the SSL model with each class and construct a classifier \( g_{\text{ssl}} \). Further details on the \( g_{\text{ssl}} \) are provided in~\cref{sec:Methodology}.

\noindent\textbf{Target Adaptation.}
We follow the CTTA protocol~\cite{Wangetal2022cotta}, which assumes a continuously changing environment without explicit domain boundaries, to assess the adaptability of the SSL model to the target domain.
The target model \( g_t \circ f_t \) is initialized from the SSL model \( g_{\text{ssl}} \circ f_{\text{ssl}} \). Our main objective is to adapt to the target domain by leveraging an online stream of unlabeled target data \( x^t \in \mathcal{X}^t \) while minimizing the mean error as the domain gradually shifts.

\subsection{Methodology}
\label{sec:Methodology}
\noindent\textbf{Prototype Classifier.}
The primary challenge in applying the SSL model to TTA is the absence of a classifier corresponding to each class. Linear probing and the $k$-nearest neighbor ($k$-NN) classifier are widely used methods for building a classifier that aligns with each class~\cite{oord2018representation,he2020momentum,chen2020simple}. However, linear probing necessitates backpropagation for gradient computation, whereas the $k$-NN classifier entails substantial computational and memory overhead due to the requirement of storing a large number of feature representations. Inspired by the prototype-based classification in few-shot learning~\cite{snell2017prototypical,2013ncm} and continual learning~\cite{rebuffi2017icarl,Hou2019lucir}, we establish a prototype $\mu_c$ for each class $c$ and employ a cosine similarity-based classifier. Using only the forward pass enhances computational efficiency. The prediction probability for each class is given by
\begin{align}
    p_t(y = c | x) = \frac{\exp(\sigma \cdot  \text{cos}(f_t(x), \mu_c))}{\sum_{i\in C} \exp(\sigma \cdot \text{cos}(f_t(x), \mu_{i}))},
\end{align}
where $cos(\cdot, \cdot)$ denotes the cosine similarity between two vectors, $\sigma$ represents the logit scaling factor, $C$ denotes the total number of classes and $\mu_c$ is the mean of features for each class $c$ for the source dataset $\{\mathcal{X}^s,\mathcal{Y}^s\}$ of the SSL model, i.e., $\mu_c = \frac{1}{|\mathcal{X}^s_c|}\sum_{\mathcal{Y}^s_c} f_{ssl}(x^s)$.

\noindent\textbf{Contrastive Learning.} Through a contrastive loss function, distance-based classifiers benefit from improved performance while enabling the gradual refinement of representations~\cite{oord2018representation,chen2020simple,cha2021co2l,wen2024provable}. Building on the need for robustness against uncertainty induced by domain shifts, we introduce an approximately correct contrastive learning method that integrates a refined segmentation of multiple prediction candidates~\cite{zhang2024candidate}.
Compensating for the low accuracy in the target domain, we identify samples sharing a pseudo label $\mathcal{T}^k$ within the top-$k$ predictions as positive samples. Conversely, when no common prediction exists among $\mathcal{T}^n$, which denotes the top-$n$ predictions with $n>k$, the sample is treated as a negative instance. For ambiguous samples that do not fit either category, contrastive loss is not applied. Accordingly, the indicator function is defined as 
\begin{align}
\label{eq:cl}
    \mathds{1}_{ij} = 
    \begin{cases} 1, & \text{if} \ \ \mathcal{T}^k_i \cap \mathcal{T}^k_j \neq \emptyset\\
    -1, & \text{if} \ \ \mathcal{T}^n_i \cap \mathcal{T}^n_j = \emptyset \ \ (n>k)\\
    0, & \text{otherwise.}\\ 
    \end{cases} 
\end{align}

We estimate the relationships among samples predicted as positive, ambiguous, or negative using the indicator function. By applying contrastive loss to these approximately correct sample relationships, we actively leverage the initial classification capability of the SSL model while ensuring stability. The approximately correct contrastive learning loss is defined as follows:
\begin{align}
\mathcal{L}_{cl} = -\sum_{i=1}^{B}\sum_{j=1}^{B} \frac{\mathds{1}_{ij}}{\sum_{j=1}^{B} \mathds{1}_{ij}} \log \frac{\exp(S^t_{ij})}{\sum_{k=1}^{B} \exp(S^t_{ik})},
\end{align}
where $S^t_{ij}$ represents the cosine similarity between $f_t(x_i)$ and $f_t(x_j)$, and $B$ denotes the batch size.

\noindent\textbf{Knowledge Distillation.} As a fundamental technique for transferring knowledge between models, knowledge distillation~\cite{hinton2015kd} has demonstrated effectiveness in various domains, including model compression~\cite{romero2014fitnets,zagoruyko2017atkd}, mitigating catastrophic forgetting~\cite{rebuffi2017icarl,Hou2019lucir}, improving zero-shot performance~\cite{vemulapalli2024knowledge,zhang2025accessing}. To preserve generalization performance and mitigate overfitting under continuous domain shifts, we transfer knowledge from the SSL model to the target model. By reducing the rotation of feature representations, we retain the knowledge embedded in the SSL model while ensuring prediction consistency in the prototype classifier, which relies on cosine similarity between feature vectors and weight vectors of the classifier. To this end, we propose a knowledge distillation loss that aligns normalized feature vectors, facilitating stable knowledge transfer while preserving the geometric structure of the feature space.
\begin{align}
\label{eq:kd}
    \mathcal{L}_{kd} &=\frac{1}{B}\sum_{i=1}^{B} 
    \| \overline{f}_t(x_i) - \overline{f}_{ssl}(x_i)\|_2,
\end{align}
where $\overline{f}(x) = \frac{f(x)}{\| f(x)\|}$ denotes normalized feature vector, and $\| \cdot \|_2$ represents the Frobenius norm.

\noindent\textbf{Mutual Learning.}
A self-supervised model demonstrates generalization performance by training on large-scale datasets, whereas a target model acquires domain-specific knowledge through adaptation. Drawing insight from studies suggesting that collaborative learning between models enhances robustness to noisy labels~\cite{coteaching,coteaching_plus,jocor}, we propose a collaborative mutual learning framework to integrate the strengths of these distinct predictive tendencies. To adapt the model to the target domain, we update the SSL model's classifier using pseudo labels generated by the target model, which maintains relatively high accuracy. This enables classifier refinement while preserving the fixed embeddings of the SSL model. Furthermore, we maximize the mutual information between predicted probability distributions to capture relational information between samples, leveraging the SSL model’s representational capabilities. The collaborative loss for mutual knowledge transfer is formulated as follows:
\begin{align}
\label{eq:ml}
\mathcal{L}_{ml} &=\frac{1}{B}\sum_{i=1}^{B} \ [ \underbrace{\mathcal{H}(p_i^{ssl},\hat{p}_i^t)}_\text{loss for SSL} + \underbrace{I(p^{t}_i,p^{ssl}_i)}_\text{loss for target}],
\end{align}
where $p_i^t$ denotes the probability obtained by applying the softmax function to \( g_t \circ f_t (x_i)\) and $\hat{p}^t_i=\text{argmax}(p^t_i)$. $I(p,q)$ represents the mutual information~\cite{ji2019iic}, and $\mathcal{H}(p,q)$ is cross entropy between two probability distributions $p$ and $q$.

The total loss function of the proposed method, which consists of approximately correct contrastive learning, knowledge distillation, and mutual learning, is formulated as follows:
\begin{align}
\label{eq:total}
\mathcal{L}_{aws} &= \mathcal{L}_{cl} + \lambda_{kd}\mathcal{L}_{kd} + \lambda_{ml}\mathcal{L}_{ml},
\end{align}
where $\lambda_{kd}$ and $\lambda_{ml}$ are hyperparameters for knowledge distillation loss and mutual loss, respectively.
 
\begin{table*}[t]
\centering

\small
\setlength\tabcolsep{2pt}
\begin{adjustbox}{width=0.99\linewidth,center=\linewidth}
\begin{tabular}{l|l|ccccccccccccccc|cc}
\toprule
\multicolumn{2}{c|}{Time} & \multicolumn{15}{l|}{$t\xrightarrow{\hspace*{13.5cm}}$}& \\ \hline
\multirow{1}{*}{Pretrained Model} & Method &
\rotatebox[origin=c]{50}{Gaussian} & \rotatebox[origin=c]{50}{shot} & \rotatebox[origin=c]{50}{impulse} & \rotatebox[origin=c]{50}{defocus} & \rotatebox[origin=c]{50}{glass} & \rotatebox[origin=c]{50}{motion} & \rotatebox[origin=c]{50}{zoom} & \rotatebox[origin=c]{50}{snow} & \rotatebox[origin=c]{50}{frost} & \rotatebox[origin=c]{50}{fog}  & \rotatebox[origin=c]{50}{brightness} & \rotatebox[origin=c]{50}{contrast} & \rotatebox[origin=c]{50}{elastic\_trans} & \rotatebox[origin=c]{50}{pixelate} & \rotatebox[origin=c]{50}{jpeg}
& Mean$\downarrow$ & Gain$\uparrow$\\\hline
&No Adapt~\cite{dosovitskiy2021vit}~\textcolor{gray}{[Baseline]}&53.0&51.8&52.1&68.5&78.8&58.5&63.3&49.9&54.2&57.7&26.4&91.4&57.5&38.0&36.2&55.8&0.0\\
&Tent~\cite{DequanWangetal2021}~\textcolor{gray}{[ICLR'21]}&52.2&48.9&49.2&65.8&73.0&54.5&58.4&44.0&47.7&50.3&23.9&72.8&55.7&34.4&33.9&51.0&+4.8\\
&CoTTA~\cite{Wangetal2022cotta}~\textcolor{gray}{[CVPR'22]}&52.9&51.6&51.4&68.3&78.1&57.1&62.0&48.2&52.7&55.3&25.9&90.0&56.4&36.4&35.2&54.8&+1.0\\
Source&SAR~\cite{niu2023sar}~\textcolor{gray}{[ICLR'23]}  &49.3&43.8&44.9&58.2&60.9&46.1&51.8&41.3&44.1&41.8&23.8&57.2&49.9&32.9&32.7&45.2&+10.6\\
pretrained&PETAL~\cite{brahma2023petal}~\textcolor{gray}{[CVPR'23]}&52.1&48.2&47.5&66.8&74.0&56.7&59.7&46.8&47.2&52.7&26.4&91.3&50.7&32.3&32.0&52.3&+3.5\\
Acc. 83.6\%&ViDA~\cite{liu2023vida}~\textcolor{gray}{[ICLR'24]}&47.7&42.5 &42.9& 52.2 & 56.9 & 45.5 & 48.9 & 38.9 & 42.7 & 40.7 & 24.3 & 52.8 & 49.1 & 33.5 & 33.1 & 43.4 & +12.4 \\
&Continual-MAE~\cite{liu2024continual}~\textcolor{gray}{[CVPR'24]} &46.3&41.9&42.5&51.4&54.9&43.3&\bf40.7&\bf34.2&\bf35.8&64.3&\bf23.4&60.3&\bf37.5&\bf29.2&31.4&42.5& +13.3\\
&COME~\cite{zhang2025come}~\textcolor{gray}{[ICLR'25]} &49.3&43.5&44.5&59.6&60.1&49.4&52.4&41.6&43.6&44.3&24.1&89.1&45.9&32.4&32.5&47.5&+8.3\\
&\cellcolor{Light}AWS~\textcolor{gray}{[Proposed]}
&\cellcolor{Light}\bf43.9&\cellcolor{Light}\bf39.6&\cellcolor{Light}\bf41.3&\cellcolor{Light}\bf48.9&\cellcolor{Light}\bf47.7&\cellcolor{Light}\bf42.2&\cellcolor{Light}42.9&\cellcolor{Light}35.8&\cellcolor{Light}37.3&\cellcolor{Light}\bf39.7&\cellcolor{Light}23.6&\cellcolor{Light}\bf49.8&\cellcolor{Light}\bf37.5&\cellcolor{Light}30.9&\cellcolor{Light}\bf30.3&\cellcolor{Light}\bf39.4&\cellcolor{Light}\bf+16.4\\
\midrule
&No Adapt~\cite{dosovitskiy2021vit}~\textcolor{gray}{[Baseline]}&85.7&83.6&85.7&68.7&86.5&73.3&73.4&64.3&64.3&61.8&38.1&79.8&65.7&55.8&50.8&69.2& 0.0\\
&Tent~\cite{DequanWangetal2021}~\textcolor{gray}{[ICLR'21]}&81.8&75.9&75.6&67.3&94.0&73.6&73.4&62.1&62.7&61.4&38.2&75.4&67.9&51.9&48.6&67.3& +1.9\\
&CoTTA~\cite{Wangetal2022cotta}~\textcolor{gray}{[CVPR'22]}&98.2&99.1&99.3&68.7&78.7&72.0&70.9&69.9&64.9&61.7&41.0&78.1&59.8&52.9&51.8&71.1&-1.9\\
\multirow{1}{*}{DINO~\cite{caron2021dino}}&SAR~\cite{niu2023sar}~\textcolor{gray}{[ICLR'23]} &81.0&73.5&73.3&68.8&91.0&73.0&72.1&61.8&62.5&61.1&38.2&74.6&67.6&51.7&48.5&66.6 & +2.6\\
Acc. 63.1\% &PETAL~\cite{brahma2023petal}~\textcolor{gray}{[CVPR'23]}&97.8&98.1&98.5&68.0&86.6&74.7&72.8&64.6&64.6&60.7&38.3&80.2&66.5&55.6&51.2&71.9&-2.7\\
&COME~\cite{zhang2025come}~\textcolor{gray}{[ICLR'25]} &85.7&83.5&85.7&68.6&86.5&73.3&73.4&64.2&64.2&61.6&38.1&80.3&65.7&56.5&51.2&69.2&+0.0\\
&\cellcolor{Light}AWS~\textcolor{gray}{[Proposed]}
&\cellcolor{Light}\bf65.9&\cellcolor{Light}\bf59.6&\cellcolor{Light}\bf60.7&\cellcolor{Light}\bf57.8&\cellcolor{Light}\bf59.3&\cellcolor{Light}\bf57.0&\cellcolor{Light}\bf52.7&\cellcolor{Light}\bf50.8&\cellcolor{Light}\bf50.9&\cellcolor{Light}\bf50.3&\cellcolor{Light}\bf37.0&\cellcolor{Light}\bf52.6&\cellcolor{Light}\bf49.6&\cellcolor{Light}\bf45.0&\cellcolor{Light}\bf45.6&\cellcolor{Light}\bf53.0&\cellcolor{Light}\bf+16.2\\
&\cellcolor{Light}AWS-FS~\textcolor{gray}{[Proposed]}
&\cellcolor{Light}66.7&\cellcolor{Light}61.0&\cellcolor{Light}63.0&\cellcolor{Light}59.1&\cellcolor{Light}61.5&\cellcolor{Light}57.9&\cellcolor{Light}53.5&\cellcolor{Light}52.3&\cellcolor{Light}52.1&\cellcolor{Light}51.2&\cellcolor{Light}39.1&\cellcolor{Light}54.3&\cellcolor{Light}50.7&\cellcolor{Light}46.3&\cellcolor{Light}47.7&\cellcolor{Light}54.4&\cellcolor{Light}+14.8\\
\midrule
&No Adapt~\cite{dosovitskiy2021vit}~\textcolor{gray}{[Baseline]}& 91.2 & 89.5 & 92.1 & 79.9 & 90.2 & 79.8 & 82.6 & 74.3 & 76.4 & 80.3 & 43.1 & 85.4 & 71.2 & 52.6 & 59.6 & 76.5 & 0.0\\
&Tent~\cite{DequanWangetal2021}~\textcolor{gray}{[ICLR'21]}&91.2&89.5&92.1&79.9&90.2&79.8&82.7&74.3&76.4&80.4&43.1&85.4&71.2&52.7&59.7&76.6&-0.1\\
&CoTTA~\cite{Wangetal2022cotta}~\textcolor{gray}{[CVPR'22]}&96.9&94.3&98.1&80.8&95.6&82.7&83.8&74.6&76.1&78.1&42.9&86.7&70.9&52.1&59.0&78.2&-1.7\\
\multirow{1}{*}{MoCo~\cite{chen2021moco}}&SAR~\cite{niu2023sar}~\textcolor{gray}{[ICLR'23]}  &91.1&89.1&91.2&79.9&90.7&78.7&82.0&72.6&73.7&78.0&41.6&85.4&68.8&51.0&57.2&75.4&+1.1\\
Acc. 60.0\% &PETAL~\cite{brahma2023petal}~\textcolor{gray}{[CVPR'23]}&96.9&94.3&98.1&80.8&95.6&82.7&83.9&74.8&76.2&77.8&42.9&86.4&71.1&51.9&59.2&78.2&-1.7\\
&COME~\cite{zhang2025come}~\textcolor{gray}{[ICLR'25]} &91.1&89.1&91.1&79.9&90.8&78.7&81.9&72.6&73.0&77.1&41.3&85.2&68.7&51.3&57.5&75.3&+1.2\\
&\cellcolor{Light}AWS~\textcolor{gray}{[Proposed]}
&\cellcolor{Light}90.2&\cellcolor{Light}84.6&\cellcolor{Light}83.7&\cellcolor{Light}\bf74.2&\cellcolor{Light}81.3&\cellcolor{Light}\bf73.7&\cellcolor{Light}\bf74.9&\cellcolor{Light}\bf66.1&\cellcolor{Light}\bf65.7&\cellcolor{Light}\bf68.7&\cellcolor{Light}\bf44.5&\cellcolor{Light}\bf73.0&\cellcolor{Light}\bf62.2&\cellcolor{Light}\bf48.8&\cellcolor{Light}\bf51.6&\cellcolor{Light}\bf69.5&\cellcolor{Light}\bf+7.0\\
&\cellcolor{Light}AWS-FS~\textcolor{gray}{[Proposed]}
&\cellcolor{Light}\bf89.4&\cellcolor{Light}\bf81.9&\cellcolor{Light}\bf81.4&\cellcolor{Light}79.9&\cellcolor{Light}\bf80.5&\cellcolor{Light}80.1&\cellcolor{Light}80.1&\cellcolor{Light}74.3&\cellcolor{Light}70.6&\cellcolor{Light}78.8&\cellcolor{Light}52.7&\cellcolor{Light}78.5&\cellcolor{Light}66.7&\cellcolor{Light}56.4&\cellcolor{Light}56.5&\cellcolor{Light}73.8&\cellcolor{Light}+2.7\\
\midrule
&No Adapt~\cite{dosovitskiy2021vit}~\textcolor{gray}{[Baseline]}& 86.1 & 84.2 & 86.9 & 69.3 & 87.6 & 74.6 & 73.3 & 62.3 & 62.5 & 60.3 & 36.1 & 78.5 & 62.2 & 48.9 & 47.2 & 68.0 &
0.0\\
&Tent~\cite{DequanWangetal2021}~\textcolor{gray}{[ICLR'21]}&86.1&84.0&87.2&68.8&88.4&71.3&71.2&60.5&61.3&60.3&36.3&79.4&63.2&47.1&48.0&67.5&+0.5\\
&CoTTA~\cite{Wangetal2022cotta}~\textcolor{gray}{[CVPR'22]}&86.1&84.3&87.0&69.3&87.6&77.3&73.3&61.8&61.9&60.0&36.1&78.0&61.9&48.4&46.7&68.0&+0.0\\
\multirow{1}{*}{iBOT~\cite{zhou2021ibot}}&SAR~\cite{niu2023sar}~\textcolor{gray}{[ICLR'23]}  &85.7&83.2&85.1&68.8&87.9&70.9&71.3&60.0&61.1&60.3&36.2&78.3&62.7&47.1&47.7&67.1&+0.9\\
Acc. 65.9\% &PETAL~\cite{brahma2023petal}~\textcolor{gray}{[CVPR'23]}&86.1&84.3&87.0&69.3&87.6&77.3&73.3&61.6&61.8&59.9&36.0&77.9&61.9&48.3&46.7&67.9&+0.1\\
&COME~\cite{zhang2025come}~\textcolor{gray}{[ICLR'25]} &86.2&84.2&87.0&69.2&87.6&74.5&73.3&62.4&62.5&60.3&36.2&78.4&66.2&48.9&47.1&68.0&+0.0\\
&\cellcolor{Light}AWS~\textcolor{gray}{[Proposed]}
&\cellcolor{Light}\bf56.4&\cellcolor{Light}\bf51.5&\cellcolor{Light}\bf53.4&\cellcolor{Light}\bf53.3&\cellcolor{Light}\bf55.0&\cellcolor{Light}\bf52.5&\cellcolor{Light}\bf48.5&\cellcolor{Light}\bf46.3&\cellcolor{Light}\bf48.1&\cellcolor{Light}\bf46.6&\cellcolor{Light}\bf34.8&\cellcolor{Light}\bf47.4&\cellcolor{Light}\bf44.6&\cellcolor{Light}\bf40.5&\cellcolor{Light}\bf42.8&\cellcolor{Light}\bf48.1&\cellcolor{Light}\bf+19.9\\
&\cellcolor{Light}AWS-FS~\textcolor{gray}{[Proposed]}
&\cellcolor{Light}58.2&\cellcolor{Light}53.3&\cellcolor{Light}55.2&\cellcolor{Light}55.6&\cellcolor{Light}56.0&\cellcolor{Light}54.3&\cellcolor{Light}50.8&\cellcolor{Light}48.7&\cellcolor{Light}49.7&\cellcolor{Light}48.4&\cellcolor{Light}36.4&\cellcolor{Light}49.8&\cellcolor{Light}45.8&\cellcolor{Light}42.3&\cellcolor{Light}44.3&\cellcolor{Light}49.9&\cellcolor{Light}+18.1\\
\bottomrule
\end{tabular}
\end{adjustbox}
\vspace{-2mm}
\caption{\textbf{Classification error rate (\%) for ImageNet-to-ImageNetC under CTTA scenario}. Mean (\%) denotes the average error rate across 15 target domains. Gain (\%) represents the improvement over the baseline. FS denotes the few-shot setup that utilizes a prototype classifier constructed with 30 samples per class. The \textbf{bold} indicates best performance.}
\label{tab:supervised_pc}
\vspace{-1mm}
\end{table*}

\begin{table*}[t]
\centering
\small
\setlength{\tabcolsep}{6pt}
\begin{adjustbox}{width=1\linewidth,center=\linewidth}
\begin{tabular}{l|cc|cc|cc|cc}
\toprule
Pretrained Model & \multicolumn{2}{c|}{Source Pretrained} & \multicolumn{2}{c|}{DINO~\cite{caron2021dino}} & \multicolumn{2}{c|}{MoCo~\cite{chen2021moco}}& \multicolumn{2}{c}{iBOT~\cite{zhou2021ibot}}\\
\midrule
Method & CIFAR10C  & CIFAR100C & CIFAR10C & CIFAR100C & CIFAR10C & CIFAR100C & CIFAR10C  & CIFAR100C\\
\midrule
No Adapt~\cite{dosovitskiy2021vit}~\textcolor{gray}{[Baseline]}   & 28.2 (0.0)          & 35.4 (0.0)     &44.3 (0.0)      &64.1 (0.0)       &42.2 (0.0)   &64.2 (0.0)  &48.0 (0.0)  &65.6 (+0.0)        \\
Tent~\cite{DequanWangetal2021}~\textcolor{gray}{[ICLR'21]}        & 23.5 (+4.7)         & 32.1 (+3.3)    &43.5 (+0.8)     &62.9 (+1.2)      &42.7 (-0.5)  &64.4 (-0.2) &45.8 (+2.2) &53.3 (+12.3)        \\
CoTTA~\cite{Wangetal2022cotta}~\textcolor{gray}{[CVPR'22]}        & 24.6 (+3.6)         & 34.8 (+0.6)    &44.3 (+0.0)     &64.1 (+3.0)      &42.2 (+0.0)  &64.3 (-0.1) &46.6 (+1.4) &65.2 (+0.4)        \\
SAR~\cite{niu2023sar}~\textcolor{gray}{[ICLR'23]}                 & 26.6 (+1.6)         & 26.2 (+9.2)    &43.2 (+1.1)     &54.9 (+9.2)      &42.2 (+0.0)  &64.2 (+0.0) &40.2 (+7.8) &51.2 (+14.4)        \\
PETAL~\cite{brahma2023petal}~\textcolor{gray}{[CVPR'23]}          & 24.4 (+3.8)         & 28.0 (+7.4)    &36.4 (+7.9)     &60.2 (+3.9)      &42.2 (+0.0)  &64.6 (-0.4)          &46.0 (+2.0) &56.3 (+9.3)        \\
ViDA~\cite{liu2023vida}~\textcolor{gray}{[ICLR'24]}               & 20.7 (+7.5)         & 27.3 (+8.1)    &N/A             &N/A              &N/A    &N/A &N/A &N/A\\
Continual-MAE~\cite{liu2024continual}~\textcolor{gray}{[CVPR'24]} & 12.6 (+15.6)        & 26.4 (+9.0)    &N/A             &N/A              &N/A    &N/A &N/A &N/A        \\
COME~\cite{zhang2025come}~\textcolor{gray}{[ICLR'25]}             & 26.6 (+1.6)         & 25.6 (+9.8)    &42.6 (+1.7)     &61.1 (+3.0)      &42.2 (+0.0) &64.2 (+0.0) &45.0 (+3.0) &60.5 (+5.1)        \\
\rowcolor{Light}AWS~\textcolor{gray}{[Proposed]}                  & \bf10.8 (+17.4)     & \bf20.4 (+15.0)&\bf26.8 (+17.5) &\bf50.6 (+13.5)  &\bf40.7 (+1.5) &\bf62.1 (+2.1) &\bf30.1 (+17.9) &\bf50.2 (+15.4)    \\
\rowcolor{Light}AWS-FS~\textcolor{gray}{[Proposed]}& N/A & N/A &28.2 (+16.1) &52.5 (+11.6) &43.9 (-1.7) &64.3 (-0.1) &31.6 (+16.4) &51.9 (+13.7)    \\
\bottomrule
\end{tabular}
\end{adjustbox}
\vspace{-2mm}
\caption{\textbf{Summary of mean classification error (\%) on CIFAR10C and CIFAR100C under CTTA}. Mean (\%) denotes the average error rate across 15 target domains. FS denotes the few-shot setup that utilizes a prototype classifier constructed with 30 samples per class.}
\vspace{-1mm}
\label{tab:summary_cifar_ctta}
\end{table*}

\section{Experiments}
In this section, we begin by evaluating the mean error on the target domain using a source pretrained model, which is pretrained on the source domain~\cite{imagenet,recht2018cifar,krizhevsky2009cifar} in a supervised manner. We then apply this evaluation to our proposed self-supervised TTA protocol, utilizing DINO~\cite{caron2021dino}, MoCo~\cite{chen2021moco}, and iBOT~\cite{zhou2021ibot} to measure the mean error.
\cref{sec:4.2} describes the results for the source pretrained models, and \cref{sec:4.3} for the self-supervised models.

\subsection{Experimental setup} 
\noindent\textbf{Datasets and Models.}
\label{exp}
We conduct our experiments on standard CTTA benchmarks, including ImageNet-to-ImageNetC~\cite{hendrycks2019benchmarking}, CIFAR10-to-CIFAR10C, and CIFAR100-to-CIFAR100C~\cite{krizhevsky2009cifar}. ImageNetC, CIFAR10C, and CIFAR100C are corruption sets for each source data, with 15 types of 4 main categories, which serve as sequential target domains. Following~\cite{Wangetal2022cotta, liu2023vida,liu2024continual}, we sequentially adapt the pretrained model to 15 target domains with the highest corruption level of 5 and evaluate its online prediction performance by measuring the mean error rate. Following~\cite{liu2023vida, liu2024continual}, we adopt ViT-B/16~\cite{dosovitskiy2021vit} as the backbone network. We present experimental results for both source pretrained and self-supervised models, using DINO~\cite{caron2021dino}, MoCo~\cite{chen2021moco}, and iBOT~\cite{zhou2021ibot} as SSL models.

\noindent\textbf{Compared Methods.} 
We compare our AWS with the well-known state-of-the-art methods: Tent~\cite{DequanWangetal2021}, SAR~\cite{niu2023sar}, COME~\cite{zhang2025come}, CoTTA~\cite{Wangetal2022cotta}, PETAL~\cite{brahma2023petal}, ViDA~\cite{liu2023vida}, and Continual-MAE~\cite{liu2024continual}. 
ViDA and Continual-MAE require additional training as they incorporate an extra adapter into the source model. This makes it challenging to apply them using self-supervised models. Therefore, we do not include their results on self-supervised models.

\noindent\textbf{Implementation Details.} 
We employ the SGD optimizer with a momentum of 0.9 for training on the target domain. The batch size is 64 for ImageNetC and 16 for CIFAR datasets. The learning rate is set to ${1e\text{-}4}$$\times \frac{\text{batch size}}{64}$ for the source pretrained models, and we select the range of [${1e\text{-}3}$, ${1e\text{-}4}$, ${1e\text{-}5}$, ${1e\text{-}6}$]$\times \frac{\text{batch size}}{64}$ for the self-supervised models.
More implementation details in \cref{hyperparameter}.

\begin{figure*}[t]
\begin{center}
\includegraphics[width=0.99\linewidth]{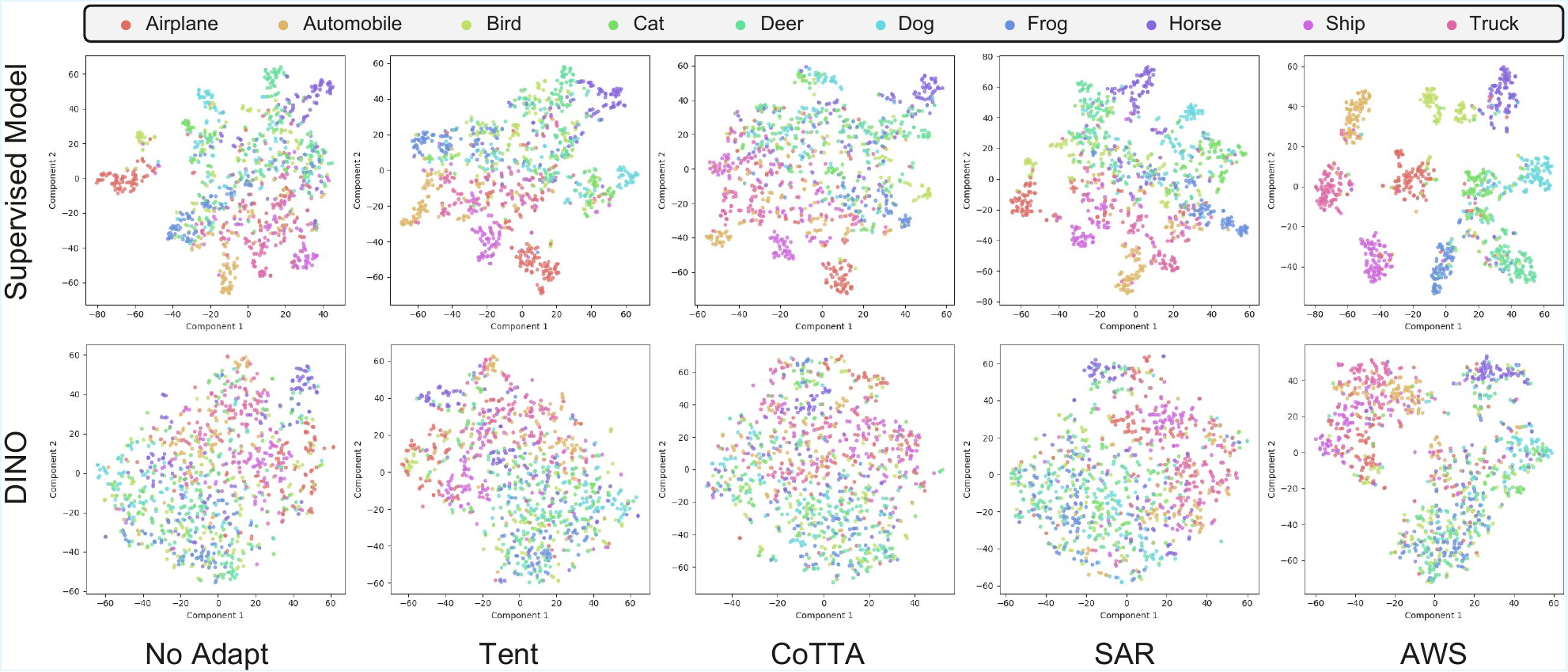}
\end{center}
\caption{\textbf{Feature visualization.} We compare the t-SNE results on CIFAR10C under Gaussian noise. (above) The results of the source pretrained model; (below) the results of the SSL model, DINO.}
\label{fig:tsne}
\end{figure*}

\subsection{Results on Source Pretrained Models} 
\label{sec:4.2}
\noindent\textbf{ImageNet-to-ImageNetC.}
~\cref{tab:supervised_pc} shows the error rate for each corruption, mean error, and the corresponding performance gain in the ImageNet-to-ImageNetC scenario. The pretrained model column indicates the source model used in the experiment and the accuracy for the source domain is under each model. As shown in the source-pretrained section, we achieve the best performance of 39.4\%, surpassing the prior state-of-the-art method, Continual-MAE.

\noindent\textbf{CIFAR10-to-CIFAR10C \& CIFAR100-to-CIFAR100C.} For the CIFAR datasets, we summarize the mean error rates in~\cref{tab:summary_cifar_ctta}. We achieve error rates of 10.8\% on CIFAR10C and 20.4\% on CIFAR100C. Compared to the prior state-of-the-art method, we observe performance gains of 2.2\% and 5.2\%, respectively. Note that we consistently improve on multiple benchmarks, which demonstrates the robustness of our approach and its adaptability to conventional TTA. Additionally, we report N/A for AWS-FS in the source pretrained setting. Since the source classifier is available, the few-shot classifier is not required.

\subsection{Results on Self-Supervised Models}
\label{sec:4.3}
\noindent\textbf{ImageNet-to-ImageNetC.}
In~\cref{tab:supervised_pc}, the experimental results on ImageNetC using each SSL model are represented below source pretrained section. The self-supervised models exhibit relatively low performance on source data compared to the source pretrained model. The absence of source-specific knowledge limits the adaptability of existing TTA methods, leading to limited performance improvements: 2.6\% with DINO, 1.2\% with MoCo, and 0.9\% with iBOT. In contrast, we observe significant improvements of 16.2\%, 7.0\%, and 19.9\% compared to the baseline. Despite limited initial performance, the proposed method improves the representation and adaptability of SSL models, suggesting its potential for online adaptation.

\noindent\textbf{CIFAR10-to-CIFAR10C \& CIFAR100-to-CIFAR100C.}
In~\cref{tab:summary_cifar_ctta}, mean error rates on CIFAR benchmarks for self-supervised models~\cite{caron2021dino, chen2021moco, zhou2021ibot} are shown to the right of the source pretrained section. We achieve 26.8\%, 40.7\%, and 30.1\% on CIFAR10C and 50.6\%, 62.1\%, and 50.2\% on CIFAR100C with DINO, MoCo, and iBOT, respectively. As mentioned in~\cref{exp}, we report N/A for adapter-based methods~\cite{liu2023vida,liu2024continual} due to the need for additional training parameters. The lower initial accuracy of MoCo probably limits its ability to adapt in the unsupervised setting. Nevertheless, AWS achieves performance gains of 1.5\% and 2.1\%, representing a meaningful improvement. In conclusion, we observe consistent performance improvements on both source-pretrained and self-supervised models, verifying the effectiveness of the proposed method on standard benchmarks.

\noindent\textbf{Few-Shot Classifier Evaluation.}
In~\cref{tab:supervised_pc,tab:summary_cifar_ctta}, we report the performance of AWS-FS, where a few-shot classifier is constructed using a subset of the source data containing 30 images per class. We achieve performance improvements of 14.8\%, 2.7\%, and 18.1\% for DINO, MoCo, and iBOT, respectively. Despite the relatively lower gains due to lower source domain performance compared to the full-shot classifiers, reported as AWS; it still achieves consistently significant improvements with respect to conventional methods.

\section{Further Analysis}
\noindent\textbf{Feature Visualization.} 
We provide t-SNE~\cite{van2008visualizing} visualization results to analyze the effect of TTA methods on the distribution of representations in~\cref{fig:tsne}. After adaptation, we extract features from the Gaussian noise corruption in CIFAR10C using both the source pretrained model and the self-supervised model, DINO. Existing approaches are typically designed to preserve the initial representations by updating only normalization layers or employing an EMA model. Consequently, these conservative update strategies demand high initial performance of the source model, leading to dependency on its initial state. In contrast, we observe that the proposed method exhibits improved decision boundaries for both the source pretrained model and the self-supervised model.

\begin{table*}[t]
\centering
\subfloat[
    \textbf{Hyperparameter $[k,n]$.}
    \label{tab:ablation_alpha}
]{
    \centering
    \begin{minipage}{0.32\linewidth}{
        \begin{center}
            \tablestyle{4pt}{1.0}
            \small
            \begin{tabular}{x{22}x{26}x{26}x{30}}
                \shline
                $[k,n]$&IN-C&C10-C&C100-C\\
                \shline
                $[1,2]$  & 39.5 & 11.5 & \cellcolor{Light}\bf20.4 \\
                $[1,3]$  & 39.5 & 11.0 & 20.5 \\
                $[1,5]$  & \cellcolor{Light}\bf39.4 & \cellcolor{Light}\bf10.8 & 20.7 \\
                $[3,10]$  & 40.1 & 57.2 & 23.1 \\
                $[5,20]$  & 40.4 & N/A & 24.8 \\
                \shline
            \end{tabular}
    \end{center}}
    \end{minipage}
}
\hspace{0.01em}
\subfloat[
    \textbf{Hyperparameter $\lambda_{kd}$.}
    \label{tab:ablation_beta}
]{
    \centering
    \begin{minipage}{0.32\linewidth}{
        \begin{center}
            \tablestyle{4pt}{1.0}
            \small
            \begin{tabular}{x{22}x{26}x{26}x{30}}
                \shline
                $\lambda_{kd}$&IN-C&C10-C&C100-C\\
                \shline
                0 & 40.6 & 11.1 & 22.3 \\
                0.01 & \cellcolor{Light}\bf39.4 & \cellcolor{Light}\bf10.8 & \cellcolor{Light}\bf20.4 \\
                0.02 & 40.1 & 11.2 & 22.5 \\
                0.03 & 41.9 & 11.7 & 24.9 \\
                0.04 & 43.6 & 11.5 & 26.9 \\
                \shline
            \end{tabular}
    \end{center}}
    \end{minipage}
}
\hspace{0.01em}
\subfloat[
    \textbf{Hyperparameter $\lambda_{ml}$.}
    \label{tab:ablation_gamma}
]{
    \centering
    \begin{minipage}{0.32\linewidth}{
        \begin{center}
            \tablestyle{4pt}{1.0}
            \small
            \begin{tabular}{x{22}x{26}x{26}x{30}}
                \shline
                $\lambda_{ml}$&IN-C&C10-C&C100-C\\
                \shline		
                0 	  & 43.8 & 13.7 & 25.1 \\
                0.1   & 41.4 & 12.8 & 23.9 \\
                0.2   & 40.3 & 11.7 & 21.5 \\
                0.3   & 39.8 & 11.6 & 20.8 \\
                0.4   & \cellcolor{Light}\bf39.4 & \cellcolor{Light}\bf10.8 & \cellcolor{Light}\bf20.4 \\
                \shline
            \end{tabular}
    \end{center}}
    \end{minipage}
}
\caption{\textbf{AWS ablation experiments.} We investigate the sensitivity of hyperparameters in proposed method. IN-C, C10-C, and C100-C are ImageNetC, CIFAR10C, and CIFAR100C, respectively.}
\label{tab:ablations}\
\end{table*}

\noindent\textbf{Hyperparameter Analysis.}
The proposed method involves four hyperparameters: $k$, $n$, $\lambda_{kd}$, and $\lambda_{ml}$. We conduct a grid search in~\cref{tab:ablations} to analyze the sensitivity across all datasets using the source pretrained model. According to~\cref{tab:ablation_alpha}, the best performing configurations of $[k,n]$ are $[1,5]$ for ImageNetC and CIFAR10C, and $[1,2]$ for CIFAR100C. Moreover, $\lambda_{kd}$ and $\lambda_{ml}$ represents that the best performance is obtained with $\lambda_{kd}=0.01$ and $\lambda_{ml}=0.4$. We observe that our method not only exhibits low sensitivity to hyperparameters but also surpasses previous methods across a wide range of hyperparameter settings.

\noindent\textbf{Domain Generalization.}
In~\cref{tab:dg}, we evaluate the domain generalization performance on ImageNetC. Following ViDA~\cite{liu2023vida}, we adapt to 10 corruption types from ImageNetC under the CTTA protocol, and subsequently evaluate performance on the 5 remaining unseen corruption types. We achieves an 8.8\% improvement over the source model and surpasses the previous state-of-the-art by 1.2\%. These results indicate that the proposed method acquires generalized knowledge and enhances representational capacity during adaptation, thereby improving performance on unseen target domains.

\begin{table}[t]
    \centering
        \setlength\tabcolsep{0.05cm}
        \small
        \setlength\tabcolsep{3pt}
        \begin{adjustbox}{width=0.98\linewidth,center=\linewidth}
        \begin{tabular}{l|ccccc|c}
        \toprule
         \multirow{2.3}{*}{Method} &  \multicolumn{5}{c|}{\textbf{Directly test on unseen domains}}& \multicolumn{1}{c}{\textbf{Unseen}} \\ 
         \cmidrule(lr){2-7}
         & bri. & contrast & elastic & pixelate & jpeg 
        & Mean$\downarrow$\\
        \midrule
        No Adapt&26.4&91.4&57.5&38.0&36.2&49.9\\
        Tent &25.8&91.9&57.0&37.2&35.7&49.5\\
       CoTTA &25.3&88.1&55.7&36.4&34.6&48.0\\
       ViDA & \bf24.6& 68.2& 49.8& 34.7& 34.1& 42.3\\
        \rowcolor{Light}AWS&24.8&\bf65.9&\bf47.1&\bf34.1&\bf33.5&\bf41.1\\
        \bottomrule
        \end{tabular}
        \end{adjustbox}
        \caption{\textbf{Domain generalization} performance on ImageNetC. Results (\%) represent the error rates for unseen domains.}
        \label{tab:dg}
\end{table}

\noindent\textbf{Effectiveness of Individual Components.}
\cref{tab:component} presents an ablation study evaluating the contribution of each component in our method, including CL, KD and ML. First, we apply CL to enhance the representational capability of the SSL model and achieve performance improvements of 12.4\%, 12.2\%, and 7.5\% on ImageNetC, CIFAR10C, and CIFAR100C, respectively (row1). These results indicate that applying CL individually can effectively improve adaptation. Second, when KD is introduced to CL, we observe that it mitigates forgetting during adaptation and results in comparable or even lower mean error than using CL alone (row4). Third, adding ML to the combination of CL and KD achieves the best performance across all datasets, demonstrating that ML provides additional benefits for further performance improvement (row6). The ablation study suggests that each component contributes to complementary aspects of the adaptation process.

\begin{table}[t]
    \small
    \centering
    \begin{adjustbox}{width=0.85\linewidth,center=\linewidth}
    \begin{tabular}{x{20}x{20}x{20}x{25}x{25}x{29}}
        \toprule
        CL & KD & ML & IN-C & C10-C & C100-C\\
        \midrule
        \rowcolor{gray! 10}\multicolumn{3}{c}{No Adapt~\textcolor{gray}{[Baseline]}}    & 55.8 & 28.2 & 35.4 \\
        \cmark & &   & 43.4 & 16.0 & 27.9 \\
        & & \cmark   & 42.7 & 21.3 & 21.8 \\
        & \cmark& \cmark     & 41.6 & 21.3 & 23.0 \\
        \cmark & \cmark &    & 43.8 & 14.1 & 25.1 \\
        \cmark & & \cmark    & 40.6 & 11.2 & 22.2 \\
        \midrule
        \rowcolor{Light}\cmark & \cmark& \cmark  & \bf39.4 & \bf10.8 & \bf20.4 \\
        \bottomrule
    \end{tabular}
    \end{adjustbox}
    \caption{\textbf{Effect of each component}, such as Contrastive Learning (CL), Knowledge Distillation (KD), and Mutual Learning (ML). 
    }
    \label{tab:component}
\end{table}

\section{Conclusion}
In this paper, we investigate the feasibility of integrating self-supervised models into TTA and explore effective strategies to enhance their adaptability within this scenario.
We address the primary challenge of applying self-supervised models to TTA, the absence of a classifier, by proposing a prototype classifier without extra training and cost.
Furthermore, we propose AWS, composed of CL, KD, and ML, to effectively leverage the expressive representations of self-supervised models while reducing reliance on source-specific knowledge for more stable adaptation.
Extensive experiments demonstrate that our proposed AWS is highly effective not only in the self-supervised setting but also in the conventional supervised setting.
Based on these results, we expect this study to contribute to expanding the potential of self-supervised models in TTA and hope that future research will build on these findings.

{
    \small
    \bibliographystyle{ieeenat_fullname}
    \bibliography{main}
}

\newpage
\appendix
\onecolumn

\section*{Appendix}
\section{Hyperparameters in experiments}
\label{hyperparameter}
In this section, we describe the learning rate configuration used in our experiments on SSL models. 
For fair comparison, we select the learning rate that yields the lowest mean error within the range defined in [${1e\text{-}3}$, ${1e\text{-}4}$, ${1e\text{-}5}$, ${1e\text{-}6}$]$\times \frac{\text{batch size}}{64}$.

\begin{enumerate}
    \item \textbf{DINO}
    \begin{itemize}
        \item  Learning rate (ImageNetC) [$\textcolor{BrickRed}{\bf1e\text{-}3}$ (Tent, SAR, CoTTA), $\textcolor{BrickRed}{\bf1e\text{-}4}$ (PETAL, AWS), $\rm1e\text{-}5$, $\textcolor{BrickRed}{\bf1e\text{-}6}$ (COME)] $\times \frac{\text{batch size}}{64}$
        \item  Learning rate (CIFAR10C) [$\textcolor{BrickRed}{\bf1e\text{-}3}$ (CoTTA, PETAL), $\textcolor{BrickRed}{\bf1e\text{-}4}$ (Tent, SAR, COME), $\textcolor{BrickRed}{\bf1e\text{-}5}$ (AWS), $\rm1e\text{-}6$] $\times \frac{\text{batch size}}{64}$
        \item  Learning rate (CIFAR100C) [$\textcolor{BrickRed}{\bf1e\text{-}3}$ (SAR, PETAL), $\textcolor{BrickRed}{\bf1e\text{-}4}$ (Tent, COME, CoTTA), $\textcolor{BrickRed}{\bf1e\text{-}5}$ (AWS), $\rm1e\text{-}6$] $\times \frac{\text{batch size}}{64}$
    \end{itemize}
    \item \textbf{MoCo}
    \begin{itemize}
        \item  Learning rate (ImageNetC) [$\textcolor{BrickRed}{\bf1e\text{-}3}$ (CoTTA, PETAL), $\textcolor{BrickRed}{\bf1e\text{-}4}$ (SAR, COME, AWS), $\rm1e\text{-}5$, $\textcolor{BrickRed}{\bf1e\text{-}6}$ (Tent)] $\times \frac{\text{batch size}}{64}$
        \item  Learning rate (CIFAR10C) [$\textcolor{BrickRed}{\bf1e\text{-}3}$ (SAR, COME, CoTTA), $\rm1e\text{-}4$, $\textcolor{BrickRed}{\bf1e\text{-}5}$(AWS), $\textcolor{BrickRed}{\bf1e\text{-}6}$ (Tent, PETAL)] $\times \frac{\text{batch size}}{64}$
        \item  Learning rate (CIFAR100C) [$\textcolor{BrickRed}{\bf1e\text{-}3}$ (CoTTA, COME, SAR), $\rm1e\text{-}4$, $\textcolor{BrickRed}{\bf1e\text{-}5}$(AWS), $\textcolor{BrickRed}{\bf1e\text{-}6}$ (Tent, PETAL)] $\times \frac{\text{batch size}}{64}$
    \end{itemize}
    \item \textbf{iBOT}
    \begin{itemize}
        \item  Learning rate (ImageNetC) [$\rm1e\text{-}3$, $\textcolor{BrickRed}{\bf1e\text{-}4}$ (Tent, SAR, CoTTA, PETAL, AWS), $\rm1e\text{-}5$, $\textcolor{BrickRed}{\bf1e\text{-}6}$ (COME)] $\times \frac{\text{batch size}}{64}$
        \item  Learning rate (CIFAR10C) [$\textcolor{BrickRed}{\bf1e\text{-}3}$ (SAR, CoTTA), $\textcolor{BrickRed}{\bf1e\text{-}4}$ (Tent, COME), $\rm1e\text{-}5$, $\textcolor{BrickRed}{\bf1e\text{-}6}$ (PETAL, AWS)] $\times \frac{\text{batch size}}{64}$
        \item  Learning rate (CIFAR100C) [$\textcolor{BrickRed}{\bf1e\text{-}3}$ (Tent, SAR, CoTTA, PETAL), $\textcolor{BrickRed}{\bf1e\text{-}4}$ (COME), $\rm1e\text{-}5$, $\textcolor{BrickRed}{\bf1e\text{-}6}$ (AWS)] $\times \frac{\text{batch size}}{64}$
    \end{itemize}
\end{enumerate}

\section{Effect of the Number of Few-Shot Samples}
\begin{wrapfigure}{r}{0.32\linewidth}
    \centering
    \vspace{-5mm}
    \includegraphics[width=0.8\linewidth]{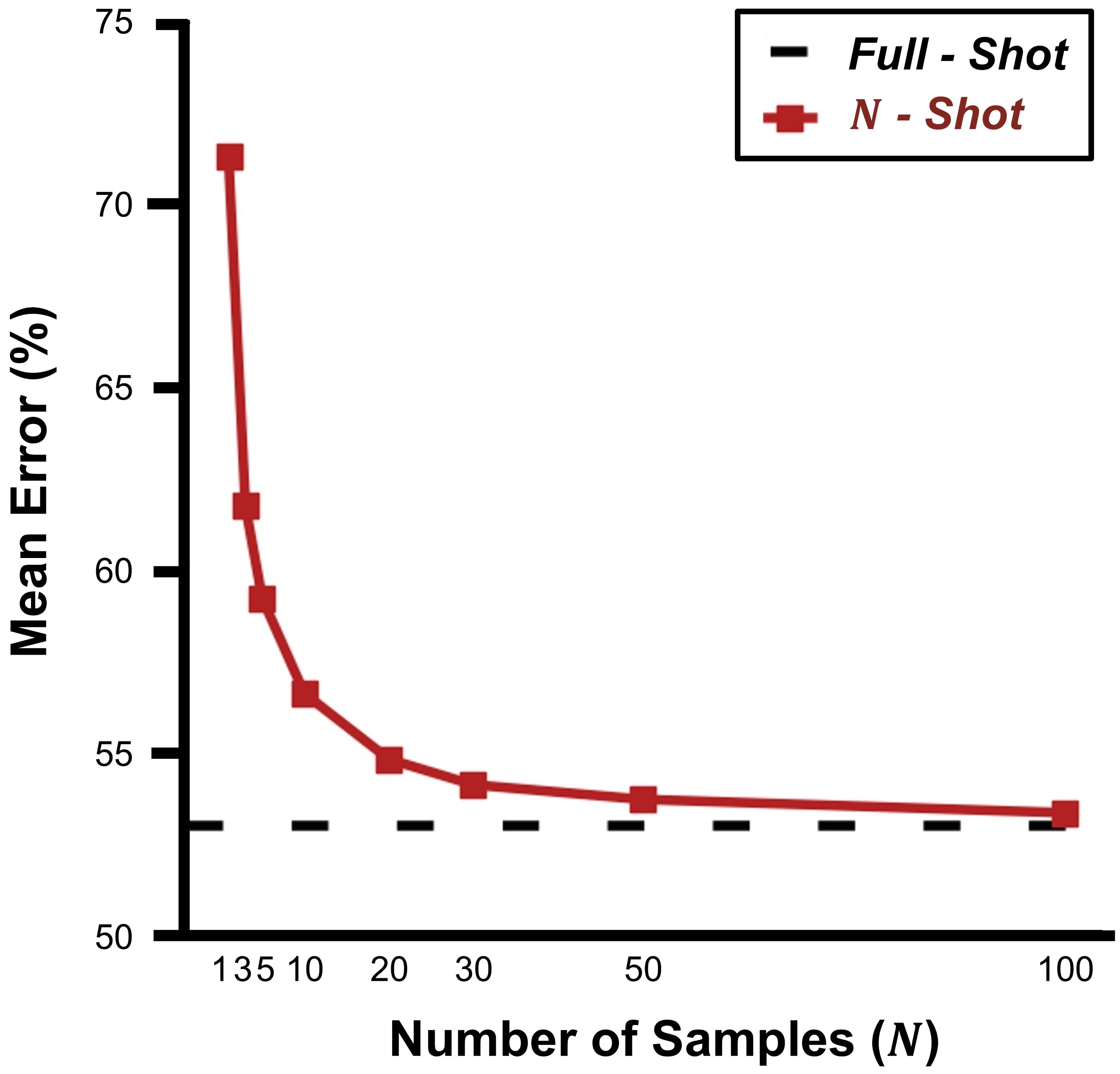}
    \vspace{-2mm}
    \caption{Effect of \# of samples}
    \label{fig:samples}
    \vspace{-5mm}
\end{wrapfigure}
We study the applicability of AWS when the classifier is constructed using a subset of the source data. To this end, we evaluate the effect of varying the number of samples ($N$) and compare it with the full-shot setting, where the entire dataset is used. Specifically, we compare the mean error rates on ImageNetC using DINO backbone across three different seeds. As shown in \cref{fig:samples}, the performance is close to that achieved using the entire dataset as $N$ increases. Notably, $N=30$ few-shot setting requires only 3\% of the source data compared to the full-shot setting. Nevertheless, the performance gap to the full-shot setting remains within 1\%. We observe that a classifier constructed with a subset of data can achieve performance close to the full-shot classifier. This observation suggests that our method has the potential to maintain its effectiveness even in more challenging scenarios.

\section{Natural Domain Shift Scenarios}
In ~\cref{tab:diverse_domain_shift}, we provide experiments on natural domain shift datasets to extend the validity of the proposed method in real-world natural changes. We conduct on ImageNet-R~\cite{hendrycks2021many}, V2~\cite{recht2019imagenet}, Sketch~\cite{wang2019learning} as a target for ImageNet~\cite{imagenet} source. We compare the results of our method with Tent~\cite{DequanWangetal2021}, CoTTA~\cite{Wangetal2022cotta}, SAR~\cite{niu2023sar} and FOA~\cite{niu2024foa}. Our method consistently achieves improved performance, demonstrating the effectiveness of AWS.

\begin{table}[hbt!]
    \centering
    \small
    \setlength{\tabcolsep}{8pt}
    \begin{adjustbox}{width=0.99\linewidth,center=\linewidth}
    \begin{tabular}{l|ccc|ccc|ccc|ccc}
        \hline
        \multirow{2.5}{*}{Method} & \multicolumn{3}{c|}{Source Pretrained} & \multicolumn{3}{c|}{DINO} & \multicolumn{3}{c|}{MoCo} & \multicolumn{3}{c}{iBOT} \\
        \cmidrule(lr){2-13}
        & R & V2 & Sketch & R & V2 & Sketch & R & V2 & Sketch & R & V2 & Sketch \\
        \hline
        No Adapt & 59.5 & \textbf{75.4} & 44.9 & 39.3 & 56.9 & 24.3 & 26.6 & 52.8 & 18.3 & 40.6 & 59.1 & 25.0 \\
        TENT & 63.9 & 75.2 & 49.1 & 39.6& 57.0 & 24.4 & 26.6& 53.2 & 18.3& 41.3& 59.1 & 25.1\\
        CoTTA & 63.5 & \textbf{75.4} & 50.0 & 39.5& 57.0 & 25.5 & 21.7& 52.1 & 11.2 & 41.1& 59.2& 26.8 \\
        SAR &63.3&75.1&48.7&39.8&57.0&24.5&\textbf{38.0}&\textbf{54.2}&19.0&41.4&59.0&25.1\\
        FOA &63.8&\textbf{75.4}&\textbf{54.4}&42.1&57.1&31.5&20.6&52.6&8.1&44.4&59.0&35.6\\
        \rowcolor{Light!70}AWS (Ours) & \textbf{69.3} & \textbf{75.4} & \textbf{54.4} & \textbf{45.0} & \textbf{57.5} & \textbf{38.0} & 35.8 & 53.5& \textbf{25.6} & \textbf{48.8}& \textbf{59.6}& \textbf{40.0} \\
        \hline
    \end{tabular}
    \end{adjustbox} \caption{\label{tab:diverse_domain_shift}Classification accuracy (\%) for ImageNet-to-ImageNet-R/V2/Sketch.}
\end{table}

\section{Comparison with Adapter-based Methods}
We present a comparison of SSL models with ViDA~\cite{liu2023vida} and Continual-MAE~\cite{liu2024continual} in~\cref{tab:adapter}. For self-supervised models, the visual domain adapters are randomly initialized.

\begin{table}[hbt!]
    \centering
    \small
    \begin{adjustbox}{width=0.65\linewidth,center=\linewidth}
    \begin{tabular}{l|cccc}
    \toprule
     Method& Source Pretrained$\downarrow$ & DINO$\downarrow$ & MoCo$\downarrow$ & iBOT$\downarrow$ \\ 
     \midrule
     ViDA & 20.7 & 43.5 & 42.1 & 47.8 \\ 
     Continual-MAE & 12.6 & 44.4 & \bf39.4 & 44.1 \\ 
     \rowcolor{Light!70}AWS (Ours) & \bf10.8 & \bf26.8 & 40.7 & \bf30.1\\
    \bottomrule
    \end{tabular}
    \end{adjustbox}
    \caption{\label{tab:adapter}Classification error rate(\%) for CIFAR10-to-CIFAR10C with adapter-based methods.}
\end{table}

\section{Parameter Update Strategy}
We use both SSL model and target model in our framework. The SSL model, trained on large-scale datasets, ensures generalization performance, whereas the target model initialized from it acquires domain-specific knowledge through adaptation. While maintaining the generalized feature representations of the SSL model, we intend to improve the classifier through pseudo labels of the target model that has relatively high accuracy. In AWS, the encoder $f_{ssl}$ is kept frozen during adaptation and the classifier $g_{ssl}$ is updated. ~\cref{framework} shows EMA updates for $f_{ssl}$ and a fixed classifier $g_{ssl}$.

\begin{table}[hbt!]
    \centering
    \small
    \begin{adjustbox}{width=0.63\linewidth}
    \begin{tabular}{l|cccc}
        \toprule
        Method & Source Pretrained$\downarrow$ & DINO$\downarrow$ & MoCo$\downarrow$ & iBOT$\downarrow$ \\ 
        \midrule
        $g_{ssl}$ (Frozen) & 39.9 & \bf53.0 & 73.0 & \bf48.1 \\ 
        $f_{ssl}$ (Update) & 40.0 & 54.5 & 70.5 & 49.7 \\ 
        \rowcolor{Light!70}AWS (Ours) & \bf39.4 & \bf53.0 & \bf69.5 & \bf48.1\\
        \bottomrule
    \end{tabular}
    \end{adjustbox}
    \captionof{table}{Effect of parameter update strategy for SSL model.}
    \label{framework}
\end{table}

\section{Prediction Shifts After Adaptation}
\begin{figure}[hbt!]
    \centering
    \includegraphics[width=0.8\linewidth]{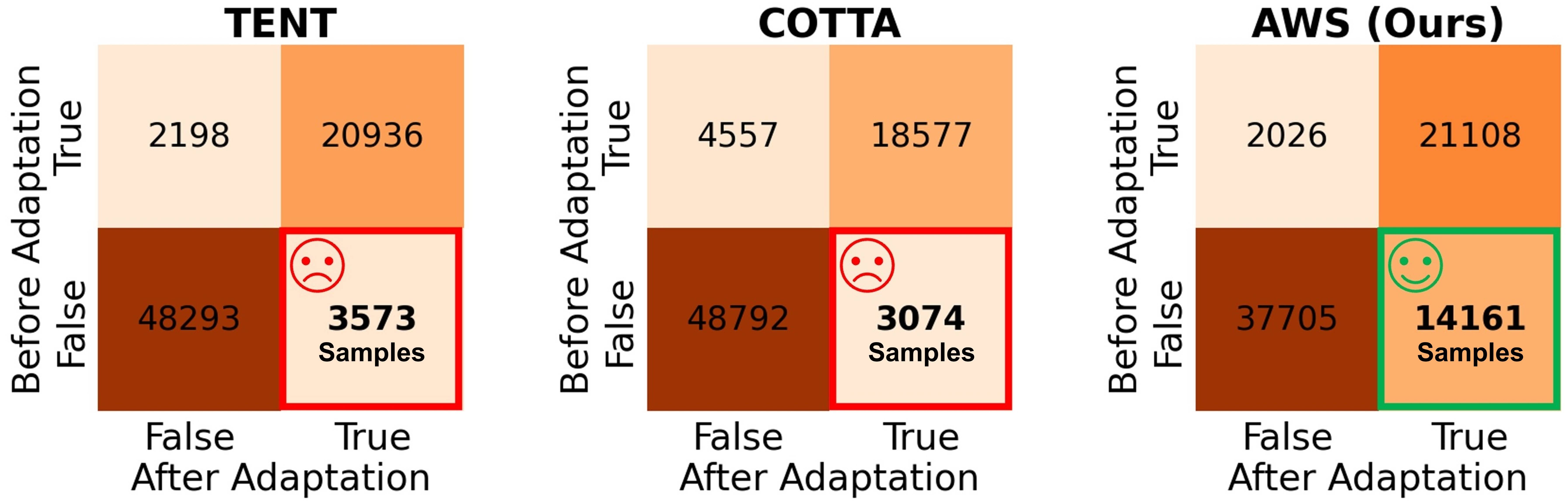}
    \caption{Prediction shifts after adaptation with respect to the source model's initial predictions.}
    \label{true_false}
\end{figure}

\cref{true_false} illustrates the change in predictions during adaptation, based on the initial predictions of the source model. The results demonstrate that our method significantly improves predictions on samples initially misclassified by the source model. We interpret this as evidence that our representation learning-based approach is less affected by confirmation bias and more effective at improving the initial model.

\section{Adaptation Time Comparison}
We compare the adaptation time for 15 domains from ImageNet-to-ImageNetC in ~\cref{adaptation_cost}. Our concern about presenting the comparison of adaptation times in the paper is that the adaptation scenarios may depend on the experimental setup, and we emphasize the efficiency of the pretraining time.

\begin{table}[hbt!]
    \centering
    \setlength\tabcolsep{0.15cm}
    \small
    \begin{adjustbox}{width=0.93\linewidth,center=\linewidth}
    \begin{tabular}{l|c|c|c}
    \toprule
     Protocol& Source Pretraining Time & Target Adaptation Time (15 domains) & Total Time$\downarrow$\\ 
     \midrule
     TTA & 1h8m23s×300epochs & 8m31s (Tent) -- 54m17s (ViDA) & $\le$ 342h9m17s \\
    \rowcolor{Light!70}SSTTA & 1m56s (Few-shot) -- 36m15s (Full) & 15m30s (Ours) & $\le$ 51m45s \\
    \bottomrule
    \end{tabular}
    \end{adjustbox}
    \caption{Comparision of adaptation cost.}
    \label{adaptation_cost}
\end{table}

\section{Full-results on CIFAR10-to-CIFAR10C and CIFAR100-to-CIFAR100C}
We provide full-results for CIFAR10-to-CIFAR10C and CIFAR100-to-CIFAR100C in~\cref{tab:supervised_cifar10c,tab:supervised_cifar100c}, respectively. Both include 15 corruption types for each source dataset. The error is measured in an online manner under sequential target domains. We conduct experiments using source pretrained model and self-supervised models such as DINO, MoCo, and iBOT.

\begin{table*}[hbt!]
\centering
\small
\setlength\tabcolsep{2pt}
\begin{adjustbox}{width=1\linewidth,center=\linewidth}
\begin{tabular}{l|l|ccccccccccccccc|cc}
\hline
\multicolumn{2}{c|}{Time} & \multicolumn{15}{l|}{$t\xrightarrow{\hspace*{13.5cm}}$}& \\ \hline
\multirow{1}{*}{PTM} & Method &
\rotatebox[origin=c]{50}{Gaussian} & \rotatebox[origin=c]{50}{shot} & \rotatebox[origin=c]{50}{impulse} & \rotatebox[origin=c]{50}{defocus} & \rotatebox[origin=c]{50}{glass} & \rotatebox[origin=c]{50}{motion} & \rotatebox[origin=c]{50}{zoom} & \rotatebox[origin=c]{50}{snow} & \rotatebox[origin=c]{50}{frost} & \rotatebox[origin=c]{50}{fog}  & \rotatebox[origin=c]{50}{brightness} & \rotatebox[origin=c]{50}{contrast} & \rotatebox[origin=c]{50}{elastic\_trans} & \rotatebox[origin=c]{50}{pixelate} & \rotatebox[origin=c]{50}{jpeg}
& Mean$\downarrow$ & Gain\\\hline
&No Adapt~\cite{dosovitskiy2021vit}&60.1&53.2&38.3&19.9&35.5&22.6&18.6&12.1&12.7&22.8&5.3&49.7&23.6&24.7&23.1&28.2&0.0\\
&Tent~\cite{DequanWangetal2021}~\textcolor{gray}{[ICLR'21]}&57.7&56.3&29.4&16.2&35.3&16.2&12.4&11.0&11.6&14.9&4.7&22.5&15.9&29.1&19.5&23.5&+4.7\\
&CoTTA~\cite{Wangetal2022cotta}~\textcolor{gray}{[CVPR'22]}&58.7&51.3&33.0&20.1&34.8&20.0&15.2&11.1&11.3&18.5&4.0&34.7&18.8&19.0&17.9&24.6&+3.6\\
Source&SAR~\cite{niu2023sar}~\textcolor{gray}{[ICLR'23]}  &54.1&47.6&38.0&19.9&34.8&22.6&18.6&12.1&12.7&22.8&5.3&39.9&23.6&24.7&23.1&26.6&+1.6\\
Pretrained&PETAL~\cite{brahma2023petal}~\textcolor{gray}{[CVPR'23]}&59.9&52.3&36.1&20.1&34.7&19.4&14.8&11.5&11.2&17.8&4.4&29.6&17.6&19.2&17.3&24.4&+3.8\\
Acc. 97.1\%&ViDA~\cite{liu2023vida}~\textcolor{gray}{[ICLR'24]}& 52.9&47.9&19.4&11.4&31.3&13.3&7.6&7.6&9.9&12.5&\bf3.8&26.3&14.4&33.9&18.2&20.7&+7.5 \\
&Continual-MAE~\cite{liu2024continual}~\textcolor{gray}{[CVPR'24]} & 30.6 & 18.9 & 11.5 & 10.4 & 22.5 & 13.9 & 9.8 & \textbf{6.6} & 6.5 & \bf8.8 & 4.0 & \bf8.5 & 12.7 & \bf9.2 & \bf14.4 &12.6& +15.6\\
&COME~\cite{zhang2025come}~\textcolor{gray}{[ICLR'25]} &54.3&47.1&36.6&19.9&34.9&22.6&18.6&12.1&12.7&22.8&5.3&40.7&23.4&24.7&23.1&26.6& +1.6\\
&\cellcolor{Light}AWS
&\cellcolor{Light}\bf18.7&\cellcolor{Light}\bf13.1&\cellcolor{Light}\bf11.0&\cellcolor{Light}\bf10.1&\cellcolor{Light}\bf18.8&\cellcolor{Light}\bf9.7&\cellcolor{Light}\bf7.2&\cellcolor{Light}7.0&\cellcolor{Light}\bf6.2&\cellcolor{Light}9.5&\cellcolor{Light}\bf3.8&\cellcolor{Light}10.2&\cellcolor{Light}\bf12.4&\cellcolor{Light}9.5&\cellcolor{Light}15.4&\cellcolor{Light}\bf10.8&\cellcolor{Light}\bf+17.4 \\
\midrule
&No Adapt~\cite{dosovitskiy2021vit}&74.8&74.5&67.2&37.0&53.7&34.7&28.8&27.7&32.5&48.6&15.3&44.2&37.8&48.9&39.8&44.3&0.0\\
&Tent~\cite{DequanWangetal2021}~\textcolor{gray}{[ICLR'21]}&76.2&77.7&68.0&36.3&54.2&34.0&27.2&26.9&30.0&45.7&13.9&38.7&35.3&51.1&36.8&43.5&+0.8\\
&CoTTA~\cite{Wangetal2022cotta}~\textcolor{gray}{[CVPR'22]}&74.8&74.5&67.1&37.1&53.7&34.7&28.9&27.6&32.4&48.5&15.3&44.1&37.7&48.8&39.7&44.3&+0.0\\
\multirow{1}{*}{DINO}&SAR~\cite{niu2023sar}~\textcolor{gray}{[ICLR'23]}  &76.7&78.1&67.4&36.4&53.6&34.0&26.9&26.2&29.0&44.7&13.7&38.1&34.9&51.8&36.2&43.2&+1.1\\
Acc. 83.1\%&PETAL~\cite{brahma2023petal}~\textcolor{gray}{[CVPR'23]}&74.6&72.1&64.2&34.4&49.1&33.2&24.5&24.1&24.0&27.7&\bf11.2&29.3&\bf19.4&36.6&\bf21.8&36.4&+7.9\\
&COME~\cite{zhang2025come}~\textcolor{gray}{[ICLR'25]}  &76.4&76.7&66.7&38.2&51.7&34.2&26.5&25.6&28.0&43.7&13.3&35.9&33.9&53.3&35.2&42.6&+1.7\\
&\cellcolor{Light}AWS
&\cellcolor{Light}\bf51.3&\cellcolor{Light}\bf33.9&\cellcolor{Light}\bf35.8&\cellcolor{Light}\bf23.3&\cellcolor{Light}\bf32.5&\cellcolor{Light}\bf26.0&\cellcolor{Light}\bf18.3&\cellcolor{Light}\bf20.4&\cellcolor{Light}\bf19.6&\cellcolor{Light}\bf26.8&\cellcolor{Light}12.6&\cellcolor{Light}\bf19.1&\cellcolor{Light}25.1&\cellcolor{Light}\bf27.4&\cellcolor{Light}29.5&\cellcolor{Light}\bf26.8&\cellcolor{Light}\bf+17.5\\
&\cellcolor{Light}AWS-FS
&\cellcolor{Light}53.2&\cellcolor{Light}35.6&\cellcolor{Light}\bf35.8&\cellcolor{Light}25.3&\cellcolor{Light}33.9&\cellcolor{Light}27.8&\cellcolor{Light}20.0&\cellcolor{Light}21.7&\cellcolor{Light}21.0&\cellcolor{Light}29.7&\cellcolor{Light}13.5&\cellcolor{Light}20.5&\cellcolor{Light}26.2&\cellcolor{Light}28.0&\cellcolor{Light}30.3&\cellcolor{Light}28.2&\cellcolor{Light}+16.1\\
\midrule
&No Adapt~\cite{dosovitskiy2021vit}& 66.7 & 66.2 & 64.7 & 36.3 & 50.8 & 39.9 & 31.5 & 25.8 & 32.7 & 55.9 & 14.0 & 29.9 & 42.3 & 45.4 & \bf31.0 & 42.2 & 0.0\\
&Tent~\cite{DequanWangetal2021}~\textcolor{gray}{[ICLR'21]}& 67.0 & 67.3 & 65.0 & 36.4 & 51.4 & 40.0 & 31.5 & 26.5 & 34.6 & 56.3 & 14.2 & 30.3 & 42.8 & 44.7 & 32.4 & 42.7 & -0.5\\
&CoTTA~\cite{Wangetal2022cotta}~\textcolor{gray}{[CVPR'22]}& 66.7 & 66.2 & 64.7 & 36.3 & 50.8 & 39.9 & 31.5 & \bf25.8 & 32.7 & 55.9 & 14.0 & 29.9 & 42.3 & 45.4 & \bf31.0 & 42.2 & +0.0\\
\multirow{1}{*}{MoCo}&SAR~\cite{niu2023sar}~\textcolor{gray}{[ICLR'23]}& 66.7 & 66.2 & 64.7 & 36.3 & 50.8 & 39.9 & 31.5 & \bf25.8 & 32.9 & 55.6 & \bf13.8 & 29.8 & 42.0 & 45.2 & 31.2 & 42.2 & +0.0\\
Acc. 83.6\%&PETAL~\cite{brahma2023petal}~\textcolor{gray}{[CVPR'23]}&66.7&66.4&64.8&36.3&51.1&39.5&30.4&26.1&33.8&54.9&14.0&29.2&41.9&44.7&33.0&42.2&+0.0\\
&COME~\cite{zhang2025come}~\textcolor{gray}{[ICLR'25]}& 66.7 & 66.2 & 64.7 & 36.3 & 50.8 & 39.9 & 31.5 & \bf25.8 & 33.0 & 55.9 & \bf13.8 & 30.6 & 42.2 & 44.9 & 31.4 & 42.2 & +0.0\\
&\cellcolor{Light}AWS
&\cellcolor{Light} \bf66.0 &\cellcolor{Light} \bf64.5 &\cellcolor{Light} \bf62.4 &\cellcolor{Light} \bf34.1 &\cellcolor{Light} \bf49.9 &\cellcolor{Light} \bf37.6 &\cellcolor{Light} \bf27.7 &\cellcolor{Light} 27.4 &\cellcolor{Light} \bf32.5 &\cellcolor{Light} \bf52.2 &\cellcolor{Light} 14.7 &\cellcolor{Light} \bf26.0 &\cellcolor{Light} \bf38.1 &\cellcolor{Light} \bf43.7 &\cellcolor{Light} 33.6 &\cellcolor{Light} \bf40.7 &\cellcolor{Light} \bf+1.5\\
&\cellcolor{Light}AWS-FS
&\cellcolor{Light}70.1&\cellcolor{Light}68.6&\cellcolor{Light}68.2&\cellcolor{Light}37.2&\cellcolor{Light}53.9&\cellcolor{Light}39.2&\cellcolor{Light}30.6&\cellcolor{Light}29.8&\cellcolor{Light}35.1&\cellcolor{Light}54.8&\cellcolor{Light}17.6&\cellcolor{Light}30.1&\cellcolor{Light}41.7&\cellcolor{Light}45.7&\cellcolor{Light}36.0&\cellcolor{Light}43.9&\cellcolor{Light}-1.7\\

\midrule
&No Adapt~\cite{dosovitskiy2021vit}& 75.8 & 75.4 & 70.2 & 51.1 & 50.1 & 43.3 & 39.5 & 25.5 & 29.3 & 54.7 & 16.9 & 48.7 & 38.8 & 59.3 & 42.2 & 48.0 & 0.0\\
&Tent~\cite{DequanWangetal2021}~\textcolor{gray}{[ICLR'21]}& 76.0 & 76.0 & 70.9 & 51.5 & 50.5 & 41.3 & 35.2 & 23.7 & 27.3 & 49.1 & 14.3 & 40.7 & 35.7 & 57.7 & 37.7 & 45.8 & +2.2\\
&CoTTA~\cite{Wangetal2022cotta}~\textcolor{gray}{[CVPR'22]}& 72.1 & 68.4 & 68.1 & 55.9 & 47.3 & 48.4 & 46.9 & 27.8 & 24.9 & 42.6 & 19.1 & 50.4 & 36.0 & 52.8 & 37.9 & 46.6 & +1.4\\
\multirow{1}{*}{iBOT}&SAR~\cite{niu2023sar}~\textcolor{gray}{[ICLR'23]}& 80.2 & 81.8 & 74.2 & 41.5 & 48.4 & 27.0 & 18.1 & 19.5 & 22.3 & 26.0 & 14.2 & 28.2 & 31.6 & 57.8 & 32.7 & 40.2 & +7.8\\
Acc. 83.4\%&PETAL~\cite{brahma2023petal}~\textcolor{gray}{[CVPR'23]}&76.7&77.0&71.7&49.8&52.1&42.0&35.9&25.2&29.5&48.3&14.8&39.4&35.6&54.1&37.6&46.0&+2.0\\
&COME~\cite{zhang2025come}~\textcolor{gray}{[ICLR'25]}& 76.4 & 76.7 & 71.5 & 52.1 & 50.8 & 38.8 & 32.3 & 23.4 & 27.6 & 45.0 & 13.4 & 37.1 & 32.9 & 60.9 & 35.5 & 45.0 & +3.0\\
&\cellcolor{Light}AWS
&\cellcolor{Light} \bf70.1 &\cellcolor{Light} \bf57.6 &\cellcolor{Light} \bf54.4 &\cellcolor{Light} \bf26.7 &\cellcolor{Light} \bf36.4 &\cellcolor{Light} \bf23.7 &\cellcolor{Light} \bf15.1 &\cellcolor{Light} \bf17.4 &\cellcolor{Light} \bf18.5 &\cellcolor{Light} \bf25.9 &\cellcolor{Light} \bf10.4 &\cellcolor{Light} \bf16.5 &\cellcolor{Light} \bf21.5 &\cellcolor{Light} 29.3 &\cellcolor{Light} \bf28.6 &\cellcolor{Light} \bf30.1 &\cellcolor{Light} \bf+17.9\\
&\cellcolor{Light}AWS-FS
&\cellcolor{Light}72.2&\cellcolor{Light}60.1&\cellcolor{Light}53.5&\cellcolor{Light}27.7&\cellcolor{Light}38.5&\cellcolor{Light}25.6&\cellcolor{Light}16.5&\cellcolor{Light}19.0&\cellcolor{Light}20.5&\cellcolor{Light}29.0&\cellcolor{Light}11.0&\cellcolor{Light}17.4&\cellcolor{Light}23.3&\cellcolor{Light}\bf29.0&\cellcolor{Light}30.6&\cellcolor{Light}31.6&\cellcolor{Light}+16.4\\
\bottomrule
\end{tabular}
\end{adjustbox}
\caption{\label{tab:supervised_cifar10c} Full-results for CIFAR10-to-CIFAR10C under CTTA scenario. Mean (\%) denotes the average error rate across 15 target domains.}
\end{table*}

\begin{table*}[hbt!]
\centering
\small
\setlength\tabcolsep{2pt}
\begin{adjustbox}{width=1\linewidth,center=\linewidth}
\begin{tabular}{l|l|ccccccccccccccc|cc}
\hline
\multicolumn{2}{c|}{Time} & \multicolumn{15}{l|}{$t\xrightarrow{\hspace*{13.5cm}}$}& \\ \hline
\multirow{1}{*}{PTM} & Method &
\rotatebox[origin=c]{50}{Gaussian} & \rotatebox[origin=c]{50}{shot} & \rotatebox[origin=c]{50}{impulse} & \rotatebox[origin=c]{50}{defocus} & \rotatebox[origin=c]{50}{glass} & \rotatebox[origin=c]{50}{motion} & \rotatebox[origin=c]{50}{zoom} & \rotatebox[origin=c]{50}{snow} & \rotatebox[origin=c]{50}{frost} & \rotatebox[origin=c]{50}{fog}  & \rotatebox[origin=c]{50}{brightness} & \rotatebox[origin=c]{50}{contrast} & \rotatebox[origin=c]{50}{elastic\_trans} & \rotatebox[origin=c]{50}{pixelate} & \rotatebox[origin=c]{50}{jpeg}
& Mean$\downarrow$ & Gain\\\hline
&No Adapt~\cite{dosovitskiy2021vit}&55.0&51.5&26.9&24.0&60.5&29.0&21.4&21.1&25.0&35.2&11.8&34.8&43.2&56.0&35.9&35.4&0.0\\
&Tent~\cite{DequanWangetal2021}~\textcolor{gray}{[ICLR'21]}&53.0&47.0&24.6&22.3&58.5&26.5&19.0&21.0&23.0&30.1&11.8&25.2&39.0&47.1&33.3&32.1&+3.3\\
&CoTTA~\cite{Wangetal2022cotta}~\textcolor{gray}{[CVPR'22]}&55.0&51.3&25.8&24.1&59.2&28.9&21.4&21.0&24.7&34.9&11.7&31.7&40.4&55.7&35.6&34.8&+0.6\\
Source&SAR~\cite{niu2023sar}~\textcolor{gray}{[ICLR'23]}  &39.4&31.0&19.8&20.9&43.9&22.6&19.1&20.3&20.2&24.3&11.8&22.3&35.2&32.1&30.1&26.2&+9.2\\
Pretrained&PETAL~\cite{brahma2023petal}~\textcolor{gray}{[CVPR'23]}&49.2&38.7&24.1&26.3&38.2&25.4&19.4&21.0&19.3&26.6&15.4&31.8&28.3&26.6&29.5&28.0&+7.4\\
Acc. 92.6\%&ViDA~\cite{liu2023vida}~\textcolor{gray}{[ICLR'24]}&50.1 & 40.7 & 22.0 & 21.2 & 45.2 & 21.6 & 16.5 & 17.9 & 16.6 & 25.6 & \bf11.5 & 29.0 & 29.6 & 34.7 & 27.1 & 27.3 & +8.1\\
&Continual-MAE~\cite{liu2024continual}~\textcolor{gray}{[CVPR'24]} & 48.6 & 30.7 & 18.5 & 21.3 & 38.4 & 22.2 & 17.5 & 19.3 & 18.0 & 24.8 & 13.1 & 27.8 & 31.4 & 35.5 & 29.5&26.4 &+9.0 \\
&COME~\cite{zhang2025come}~\textcolor{gray}{[ICLR'25]}&39.5&30.5&19.7&20.7&41.8&22.5&17.2&20.2&17.3&23.7&12.8&22.3&34.7&32.2&29.6&25.6&+9.8\\
&\cellcolor{Light}AWS
&\cellcolor{Light}\bf29.0&\cellcolor{Light}\bf24.0&\cellcolor{Light}\bf17.2&\cellcolor{Light}\bf17.8&\cellcolor{Light}\bf30.5&\cellcolor{Light}\bf19.3&\cellcolor{Light}\bf15.7&\cellcolor{Light}\bf16.7&\cellcolor{Light}\bf15.7&\cellcolor{Light}\bf19.2&\cellcolor{Light}11.8&\cellcolor{Light}\bf15.9&\cellcolor{Light}\bf25.9&\cellcolor{Light}\bf20.6&\cellcolor{Light}\bf27.0&\cellcolor{Light}\bf20.4&\cellcolor{Light}\bf+15.0 \\
\midrule
&No Adapt~\cite{dosovitskiy2021vit}&82.1&80.6&78.4&57.9&78.2&55.7&49.5&52.0&55.4&69.4&36.3&66.4&62.6&72.6&64.6&64.1&0.0\\
&Tent~\cite{DequanWangetal2021}~\textcolor{gray}{[ICLR'21]}&80.8&78.8&79.0&58.3&78.1&55.1&47.9&50.0&52.1&67.3&34.2&63.2&60.6&76.0&62.0&62.9&+1.2\\
&CoTTA~\cite{Wangetal2022cotta}~\textcolor{gray}{[CVPR'22]}&82.1&80.6&79.0&57.9&78.3&55.6&49.5&51.9&55.2&69.3&36.4&66.4&62.6&72.7&64.5&64.1&+0.0\\
\multirow{1}{*}{DINO}&SAR~\cite{niu2023sar}~\textcolor{gray}{[ICLR'23]}  &75.3&66.3&71.0&56.1&71.2&51.6&\bf41.2&\bf41.7&\bf42.7&\bf49.9&\bf32.1&48.0&\bf49.5&75.0&\bf51.4&54.9&+9.2\\
Acc. 61.5\%&PETAL~\cite{brahma2023petal}~\textcolor{gray}{[CVPR'23]}&81.7&79.0&78.0&57.2&70.5&54.1&48.1&50.9&46.6&55.0&41.2&63.1&51.4&69.5&56.2&60.2&+3.9\\
&COME~\cite{zhang2025come}~\textcolor{gray}{[ICLR'25]}  &79.4&75.9&76.4&62.9&75.0&54.8&47.9&46.9&49.0&62.5&33.2&58.3&58.8&76.8&58.2&61.1&+3.0\\

&\cellcolor{Light}AWS
&\cellcolor{Light}\bf63.7&\cellcolor{Light}\bf52.5&\cellcolor{Light}\bf58.8&\cellcolor{Light}\bf51.2&\cellcolor{Light}\bf59.7&\cellcolor{Light}\bf51.2&\cellcolor{Light}43.9&\cellcolor{Light}44.7&\cellcolor{Light}43.2&\cellcolor{Light}51.6&\cellcolor{Light}37.5&\cellcolor{Light}\bf45.7&\cellcolor{Light}52.7&\cellcolor{Light}\bf49.9&\cellcolor{Light}52.9&\cellcolor{Light}\bf50.6&\cellcolor{Light}\bf+13.5 \\
&\cellcolor{Light}AWS-FS
&\cellcolor{Light}66.2&\cellcolor{Light}54.8&\cellcolor{Light}60.2&\cellcolor{Light}52.6&\cellcolor{Light}62.0&\cellcolor{Light}52.9&\cellcolor{Light}45.6&\cellcolor{Light}47.0&\cellcolor{Light}45.3&\cellcolor{Light}53.6&\cellcolor{Light}39.0&\cellcolor{Light}47.9&\cellcolor{Light}53.9&\cellcolor{Light}52.0&\cellcolor{Light}54.6&\cellcolor{Light}52.5&\cellcolor{Light}+11.6\\
\midrule
&No Adapt~\cite{dosovitskiy2021vit}& 82.8 & 81.6 & 83.3 & 58.5 & 73.0 & 60.0 & 51.7 & 52.5 & 56.0 & 75.0 & 37.4 & 56.7 & 65.6 & 69.3 & 59.0 & 64.2 & 0.0\\
&Tent~\cite{DequanWangetal2021}~\textcolor{gray}{[ICLR'21]}& 82.9 & 81.8 & 83.4 & 58.5 & 73.3 & 60 & 51.8 & 53.0 & 56.9 & 75.2 & 37.5 & 56.7 & 66.1 & 69.1 & 59.5 & 64.4 & -0.2\\
&CoTTA~\cite{Wangetal2022cotta}~\textcolor{gray}{[CVPR'22]}& 82.8 & 81.6 & 91.6 & 57.6 & 72.8 & 58.9 & 50.8 & 52 & 55.5 & 74.5 & \bf37.1 & 55.9 & 65.0 & 69.0 & 59.9 & 64.3 & -0.1\\
\multirow{1}{*}{MoCo}&SAR~\cite{niu2023sar}~\textcolor{gray}{[ICLR'23]}& 82.8 & 82.6 & 83.3 & 58.5 & 73.0 & 60.0 & 51.7 & 52.5 & 56.0 & 74.8 & 37.3 & 57.5 & 65.6 & 69.1 & 59.1 & 64.2 & +0.0\\
Acc. 59.5\%&PETAL~\cite{brahma2023petal}~\textcolor{gray}{[CVPR'23]}&82.9&81.7&91.1&58.5&73.2&59.7&51.6&52.6&56.2&74.6&37.2&56.2&65.5&68.7&59.5&64.6&-0.4\\
&COME~\cite{zhang2025come}~\textcolor{gray}{[ICLR'25]}& 82.8 & 81.6 & 83.3 & 58.5 & 73.0 & 60.0 & 51.7 & 52.5 & 55.9 & 74.8 & 37.3 & 57.5 & 65.6 & 69.1 & 59.0 & 64.2 & +0.0\\
&\cellcolor{Light}AWS
&\cellcolor{Light} \bf82.2 &\cellcolor{Light} \bf79.6 &\cellcolor{Light} \bf80.4 &\cellcolor{Light} \bf56.8 &\cellcolor{Light} \bf71.5 &\cellcolor{Light} \bf57.9 &\cellcolor{Light} \bf49.6 &\cellcolor{Light} \bf51.2 &\cellcolor{Light} \bf52.7 &\cellcolor{Light} \bf70.1 &\cellcolor{Light} 38.1 &\cellcolor{Light} \bf53.9 &\cellcolor{Light} \bf62.5 &\cellcolor{Light} \bf65.9 &\cellcolor{Light} \bf58.7 &\cellcolor{Light} \bf62.1 &\cellcolor{Light} \bf+2.1\\
&\cellcolor{Light}AWS-FS
&\cellcolor{Light}83.6&\cellcolor{Light}81.6&\cellcolor{Light}82.3&\cellcolor{Light}59.5&\cellcolor{Light}72.8&\cellcolor{Light}60.3&\cellcolor{Light}52.2&\cellcolor{Light}54.6&\cellcolor{Light}56.2&\cellcolor{Light}71.5&\cellcolor{Light}41.1&\cellcolor{Light}55.1&\cellcolor{Light}65.3&\cellcolor{Light}67.7&\cellcolor{Light}61.1&\cellcolor{Light}64.3&\cellcolor{Light}-0.1\\
\midrule
&No Adapt~\cite{dosovitskiy2021vit}& 81.3 & 80.3 & 81.4 & 69.2 & 70.7 & 62.1 & 57.1 & 47.4 & 48.9 & 70.5 & 37.3 & 71.5 & 61.4 & 79.6 & 66.2 & 65.6 & 0.0\\
&Tent~\cite{DequanWangetal2021}~\textcolor{gray}{[ICLR'21]}& 78.8 & 74.3 & 76.8 & 57.7 & 64.4 & 45.5 & 38.2 & 38.0 & 38.5 & 45.9 & 29.7 & 42.9 & 49.5 & 71.3 & 47.6 & 53.3 & +12.3\\
&CoTTA~\cite{Wangetal2022cotta}~\textcolor{gray}{[CVPR'22]}& 78.9 & 74.4 & 78.0 & 69.3 & 67.3 & 62.8 & 59.5 & 47.0 & 45.4 & 66.2 & 42.0 & 76.9 & 58.3 & 84.3 & 67.3 & 65.2 & +0.4\\
\multirow{1}{*}{iBOT}&SAR~\cite{niu2023sar}~\textcolor{gray}{[ICLR'23]}& \bf74.6 & 65.2 & 73.2 & 55.7 & 62.1 & \bf44.8 & 38.3 & \bf36.9 & \bf36.8 & \bf44.9 & \bf29.5 & \bf42.5 & \bf48.1 & 69.4 & \bf45.6 & 51.2 & +14.4\\
Acc. 61.0\%&PETAL~\cite{brahma2023petal}~\textcolor{gray}{[CVPR'23]}&76.5&67.4&71.6&57.3&\bf59.6&52.2&45.1&47.4&44.7&54.1&39.5&71.6&48.5&55.0&54.4&56.3&+9.3\\
&COME~\cite{zhang2025come}~\textcolor{gray}{[ICLR'25]} & 79.7 & 76.2 & 78.8 & 66.4 & 67.7 & 56.8 & 52.1 & 41.6 & 41.3 & 56.4 & 32.5 & 62.0 & 54.7 & 80.5 & 61.3 & 60.5 & +5.1\\
&\cellcolor{Light}AWS
&\cellcolor{Light} 75.6 &\cellcolor{Light} \bf64.9 &\cellcolor{Light} \bf67.0 &\cellcolor{Light} \bf47.6 &\cellcolor{Light} 60.5 &\cellcolor{Light} 45.9 &\cellcolor{Light} \bf37.8 &\cellcolor{Light} 39.9&\cellcolor{Light}39.2 &\cellcolor{Light} 47.7 &\cellcolor{Light} 31.5 &\cellcolor{Light} 44.5 &\cellcolor{Light} 48.2 &\cellcolor{Light} \bf48.5 &\cellcolor{Light} 51.7 &\cellcolor{Light} \bf50.2&\cellcolor{Light} \bf+15.4\\
&\cellcolor{Light}AWS-FS
&\cellcolor{Light}76.3&\cellcolor{Light}67.4&\cellcolor{Light}69.3&\cellcolor{Light}49.6&\cellcolor{Light}62.5&\cellcolor{Light}47.2&\cellcolor{Light}39.4&\cellcolor{Light}42.3&\cellcolor{Light}41.2&\cellcolor{Light}49.6&\cellcolor{Light}33.7&\cellcolor{Light}45.6&\cellcolor{Light}50.1&\cellcolor{Light}50.3&\cellcolor{Light}53.4&\cellcolor{Light}51.9&\cellcolor{Light}+13.7\\
\bottomrule
\end{tabular}
\end{adjustbox}
\caption{\label{tab:supervised_cifar100c}Full-results for CIFAR100-to-CIFAR100C under CTTA scenario. Mean (\%) denotes the mean error rate across 15 target domains.}
\end{table*}

\end{document}